\documentclass[journal]{IEEEtran}
\usepackage{amsmath,amsfonts}
\usepackage{algorithmicx}
\usepackage{algorithm}
\usepackage{array}
\usepackage[caption=false,font=normalsize,labelfont=sf,textfont=sf]{subfig}
\usepackage{textcomp}
\usepackage{stfloats}
\usepackage{url}
\usepackage{verbatim}
\usepackage{graphicx}
\usepackage{cite}
\usepackage[switch]{lineno}
\usepackage{hyperref}
\usepackage{color}
\usepackage{xcolor}
\usepackage{graphicx}
\usepackage{booktabs}
\usepackage{amsmath, amsthm, amsfonts, amssymb}
\usepackage{mathrsfs}
\usepackage{textcomp}
\usepackage{epstopdf}
\usepackage{multirow}
\usepackage{wrapfig}
\usepackage{subfig}
\usepackage{booktabs} % For formal tables

\usepackage{algpseudocode}
\usepackage{ragged2e}
\usepackage[normalem]{ulem}
\usepackage{cleveref}
\usepackage{bm}

\definecolor{green}{rgb}{0, 0.5, 0}
\definecolor{orange}{rgb}{0.6, 0.3, 0.1}
\definecolor{red}{rgb}{1.0, 0.0, 0.0}
\definecolor{teal}{rgb}{0.0, 0.4, 0.4}
\definecolor{purple}{rgb}{0.65,0,0.65}
\definecolor{saffron}{rgb}{0.95,0.75,0.2}
\definecolor{turquoise}{rgb}{0.0,0.5,0.5}
\definecolor{brown}{rgb}{0.5, 0.16, 0.16}
\definecolor{brickred}{rgb}{.6, .2 .1}
\definecolor{coral}{rgb}{1,0.45,0.33}
\definecolor{newcolor}{rgb}{.8,.349,.1}
\definecolor{mygreen}{RGB}{15, 153, 5}
\definecolor{myorange}{RGB}{255, 153, 5}

\usepackage{tikz,color,hyperref}

\definecolor{lime}{HTML}{A6CE39}
\DeclareRobustCommand{\orcidicon}{
\begin{tikzpicture}
\draw[lime, fill=lime] (0,0)
circle[radius=0.16]
node[white]{{\fontfamily{qag}\selectfont \tiny \.{I}D}}; 
\end{tikzpicture}
\hspace{-2mm}
}
\foreach \x in {A, ..., Z}{%
\expandafter\xdef\csname orcid\x\endcsname{\noexpand\href{https://orcid.org/\csname orcidauthor\x\endcsname}{\noexpand\orcidicon}}
}

\begin{document}
% \linenumbers
\title{One Shot Learning for Edge Detection on Point Clouds}
\newcommand{\orcidauthorA}{0009-0003-9052-6689}
\newcommand{\orcidauthorB}{0000-0002-2469-4946}

\newcommand{\orcidauthorC}{0009-0006-6671-4614}
\newcommand{\orcidauthorD}{0000-0001-6218-5715}
\newcommand{\orcidauthorE}{0000-0001-6777-7445}

\author{Zhikun Tu\hspace{-1.5mm}\orcidA{}\hspace{-1mm},
        Yuhe Zhang*\hspace{-1.5mm}\orcidB{}\hspace{-1mm}, Member, IEEE,
        Yiou Jia\hspace{-1.5mm}\orcidC{}\hspace{-1mm},
        Kang Li\hspace{-1.5mm}\orcidD{}\hspace{-1mm},
        and Daniel Cohen-Or\hspace{-1.5mm}\orcidE{}\hspace{-1mm}

\thanks{Zhikun Tu, Yuhe Zhang, Yiou Jia and Kang Li are with the School of Information Science and Technology, Northwest University, Xi'an, China. Daniel Cohen-Or is with the Department of Computer Science, Tel Aviv University, Tel Aviv, Israel. Yuhe Zhang is the corresponding author. E-mail: zhangyuhe0601@nwu.edu.cn. }% <-this % stops a space

\thanks{Manuscript submitted Aug 4, 2024}}

\markboth{IEEE TRANSACTIONS ON VISUALIZATION AND COMPUTER GRAPHICS}%
{Shell \MakeLowercase{\textit{et al.}}: A Sample Article Using IEEEtran.cls for IEEE Journals}


\maketitle
\begin{abstract} 
  
Each scanner possesses its unique characteristics and exhibits its distinct sampling error distribution. Training a network on a dataset that includes data collected from different scanners is less effective than training it on data specific to a single scanner. Therefore, we present a novel one-shot learning method allowing for edge extraction on point clouds, by learning the specific data distribution of the target point cloud, and thus achieve superior results compared to networks that were trained on general data distributions. More specifically, we present how to train a lightweight network named OSFENet (\textbf{O}ne-\textbf{S}hot \textbf{F}eature \textbf{E}xtraction \textbf{Net}work), by designing a filtered-KNN-based surface patch representation that supports a one-shot learning framework. Additionally, we introduce an RBF\_DoS module, which integrates \textbf{R}adial \textbf{B}asis \textbf{F}unction-based \textbf{D}escriptor \textbf{o}f the \textbf{S}urface patch, highly beneficial for the edge extraction on point clouds. The advantage of the proposed OSFENet is demonstrated through comparative analyses against 7 baselines on the ABC dataset, and its practical utility is validated by results across diverse real-scanned datasets, including indoor scenes like S3DIS dataset, and outdoor scenes such as the Semantic3D dataset and UrbanBIS dataset.
  
\end{abstract}

\section{Introduction}

\IEEEPARstart{P}{oint} clouds serve as common representations of physical objects and scenes, often acquired through various 3D scanning techniques. However, these techniques inherently introduce sampling errors, leading to noise in the scanned data. Edges on point clouds contain vital information about the entire shape, local structure, and silhouette of smooth surfaces, making edge detection a crucial component in numerous point cloud processing tasks. Noise in scanned data can manifest as outliers, irregularities, or inaccuracies in the point cloud representation, underscoring the importance of managing and reducing noise to ensure accurate and reliable results. Despite advancements, edge detection on point clouds remains a challenging problem.

While learning-based methods such as EC-Net\cite{Yu2018ECNet}, DEF\cite{Matveev2022DEF} and NerVE\cite{Nerve2023CVPR}
have demonstrated superior performance over traditional techniques, they often rely on object-level models, each with its own unique characteristics and sampling error distribution. This variability poses challenges when training neural networks, as datasets compiled from multiple scanners may not fully capture the nuances of any individual scanner's data. 

\begin{figure}
  \includegraphics[width=0.5\textwidth]{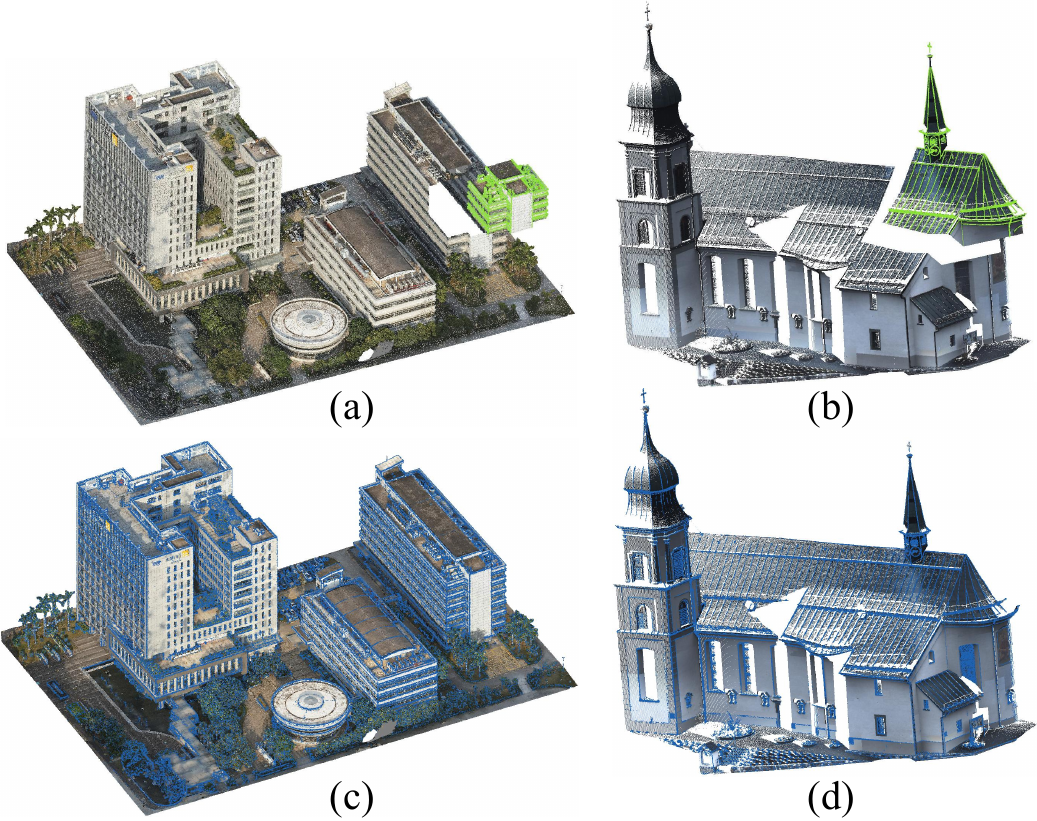}
\caption{(a) One model from UrbanBIS dataset\cite{Yang2023UrbanBISAL} with labeled edges (green). (b) One model from Semantic3D dataset\cite{Hackel2017Semantic3DnetAN} with labeled edges (green). (c) and (d): Edge extraction for the entire large scene can be achieved using our one-shot learning method  (blue indicates the predicted edges).}
  \label{teaserfigure}
\end{figure}

% Furthermore, when training and testing on a specific dataset, less emphasis has been placed on feature extraction from large-scale scene datasets. The challenges of learning-based approaches dealing with large-scale scenes mainly lie in: i) annotating large-scale point data poses a significant difficulty; for instance, a scene from the UrbanBIS dataset\textcolor{red}{[ref]} may comprise over ten million points; ii) limited availability of supervised signals for learning; for instance, the Semantic-8 dataset\textcolor{red}{[ref]} contains only 30 scenes, and the UrbanBIS dataset\textcolor{red}{[ref]} contains a similarly restricted number of scenes; iii) lengthy training processes, particularly for very large-scale models. 

In this work, we propose a novel approach that addresses these challenges by advocating for one-shot learning. We present a learning method specifically designed for edge extraction on point clouds, utilizing annotated data from a single input dataset acquired by a particular technique or scanner. By focusing on data from a specific scanner, our approach aims to mitigate the variability introduced by different scanning devices, ultimately improving the accuracy and robustness of edge detection in point clouds. More specifically, we present a one-shot learning-based edge detection technique that trains on a point cloud model or a segment of the input scanned scene. The key idea is to learn the specific data distribution from the target point cloud, and thus achieve superior results than networks that were trained on general data distributions. 

We designed a lightweight network, named OSFENet (\textbf{O}ne-\textbf{s}hot \textbf{F}eature \textbf{E}xtraction \textbf{net}work). The network is based on the observation that a point can be defined as an edge point by the distribution of its neighboring points, rather than the entire shape of the point cloud. Therefore, our edge extraction is formulated as a binary classification problem wherein the task is to categorize a point $p$ as either an edge or a non-edge. A point $p$ is selected along with its filtered-\textit{k} Nearest Neighbors (filtered-\textit{k}NN), collectively constituting the surface patch surrounding $p$, which serves as the input of OSFENet. This localized representation substantially reduces the magnitude of required labeled training data since a point cloud comprising \textit{N} points can yield \textit{N} surface patches, thereby constituting the training set.

Moreover, as demonstrated in prior studies \cite{RBFSurfaceGre, RBFSurfaceDerivatives}, the utilization of Radial Basis Functions (RBFs) for describing the surfaces has been proven to be both remarkably accurate and computationally efficient. Therefore, by integrating RBF-based surface descriptor, we develop a module named RBF\_DoS to describe the surface patch around the target point. This integration significantly enhances the performance of the proposed network, as will be validated in the ablation study. 

Both quantitative and qualitative evaluations reveal that our method outperforms selected baseline methods across five evaluation metrics on the ABC dataset\cite{Koch2019ABC}. Moreover, the qualitative results obtained from diverse real-scanned datasets ranging from indoor scenes to outdoor scenes also highlight the utility and efficiency of our proposed approach. 

Our main contributions can be summarized as follows: i) A novel one-shot learning network, which can achieve state-of-the-art edge extraction performance on point clouds, is proposed. ii) Our work represents a pioneering learning-based approach for extracting edges on large-scale scenes with limited scene data. iii) The presentation of the RBF\_DoS module which integrates RBFs-based surface descriptors in a neural network, highly promoting the performance of the network. 

\section{Related Work}
\label{relatedworks}
\subsection{Non-learning-based methods}

Before the arrival of deep-learning-based methods, the traditional methods mostly relied on: i) the distribution of points, ii) geometric invariants, iii) surface fitting or mesh reconstruction, and iv) 2D images mapped from point clouds. 

Specifically, both Gumhold et al.\cite{Gumhold2001Feature} and  Demarsin et al.\cite{Demarsin2007Detection} performed principle component analysis (PCA) on the coordinates of points to detect edge points. Similarly, Bazazian et al.\cite{Bazazian2015Fast} also employed PCA and clustering analysis for edge detection. In the geometric invariant-based method, curvature-based method \cite{Pauly2003Multi} and normal-based method \cite{Zhang2016Statistical, Weber2010Sharp} are the most typical. The success of edge detection on meshes led researchers to convert point clouds into meshes or fit local surfaces for edge detection, for example, Chen et al.\cite{Chen2022Multiscale} used the plane fitting residual to compute candidate geometric feature points; Guo et al.\cite{Guo2022SGLBP} present SGLBP which also constructed local meshes in neighborhoods for detecting edge features. 
2D image-based edge detection methods like \cite{Lin2015Line} suffer from redundant pre-processing and post-processing, gaining lower popularity.

\subsection{Deep-learning-based methods}

Owing to the release of the ABC dataset\cite{Koch2019ABC}, quite a lot of deep-learning-based methods have been done around edge detection. Yu et al.\cite{Yu2018ECNet} proposed a network named EC-Net to fit edges in the consolidation work, similarly, PIE-Net\cite{Wang2020PieNet} regards edge detection as a parametric curve inference problem which outputs parametric curves. Edges can be considered as the boundaries of segments, therefore, Loizou et al.\cite{Loizou2020Learning} used a graph convolutional network to learn part boundaries, and in JSENet\cite{Hu2020JSENet} and EDCNet \cite{Bazazian2021EDC}, edge detection and semantic segmentation were jointly performed. Li et al.\cite{primitiveSig2023} proposed a novel surface and edge detection network SED-Net for fitting geometric surfaces and edges of point clouds, which also regards edges as boundaries of geometric primitives\cite{primitiveSig2023}. 

It is proved that enriching inputs can improve the edge detection quality, for example, PCEDNet\cite{Himeur2021PCEDNet} and BoundED\cite{Bode2022BoundED} build inputs by integrating first and second-order statistics to distinguish feature points from non-feature points. Another method, DEF\cite{Matveev2022DEF}, was introduced to regress a scalar field, which represents the distance from point samples to the closest feature line on local patches. Most recently, new networks have advanced the state-of-the-art by designing new representations of point clouds, for example, Feng et al.\cite{Feng2023Deep} encodes the features using the structure of U-net and detects local information using a convolution operation; NEF\cite{NEF2023CVPR} learns a neural implicit field representing the density distribution of 3D edges; and NerVE\cite{Nerve2023CVPR} extracts piece-wise linear (PWL) curve from point clouds by learning a neural volumetric edge representation. Leveraging prior geometric insights, MSL-Net \cite{MSLNet} is crafted through the integration of an intrinsic neighbor shape descriptor, augmented normal extraction, and a cosine-based field estimation function. 

\subsection{RBF on point clouds}
Leveraging the mesh-free nature of Radial Basis Functions (RBFs) to represent the geometries can provide high accuracy and flexibility. Thus surface reconstruction from point clouds is one of the primary applications of Radial Basis Functions (RBFs)\cite{RBFSurfaceReconstruction}. Building upon RBF-based surface reconstruction, methods for approximating surface derivatives\cite{RBFSurfaceDerivatives} using RBFs have been proposed and validated for their efficacy and efficiency. In deep learning methods for point cloud processing, RBFs are primarily used to produce a smooth distribution of points within a voxel\cite{RPSNetIS2023, PerfectMatch}, which is significantly different from our method.

\subsection{Summary}
The key distinctions of our approach are: i) It learns the specific data distribution of target point clouds, enabling effective one-shot learning on datasets with limited samples, such as the 30-scene Semantic3D dataset \cite{Hackel2017Semantic3DnetAN}. ii) Our one-shot learning framework supports feature extraction on large-scale models, including outdoor scenes with over ten million points, achieving superior performance using a labeled segment of the scene.
\begin{figure*}[h]
  \centering
  \includegraphics[width=1.0 \linewidth]{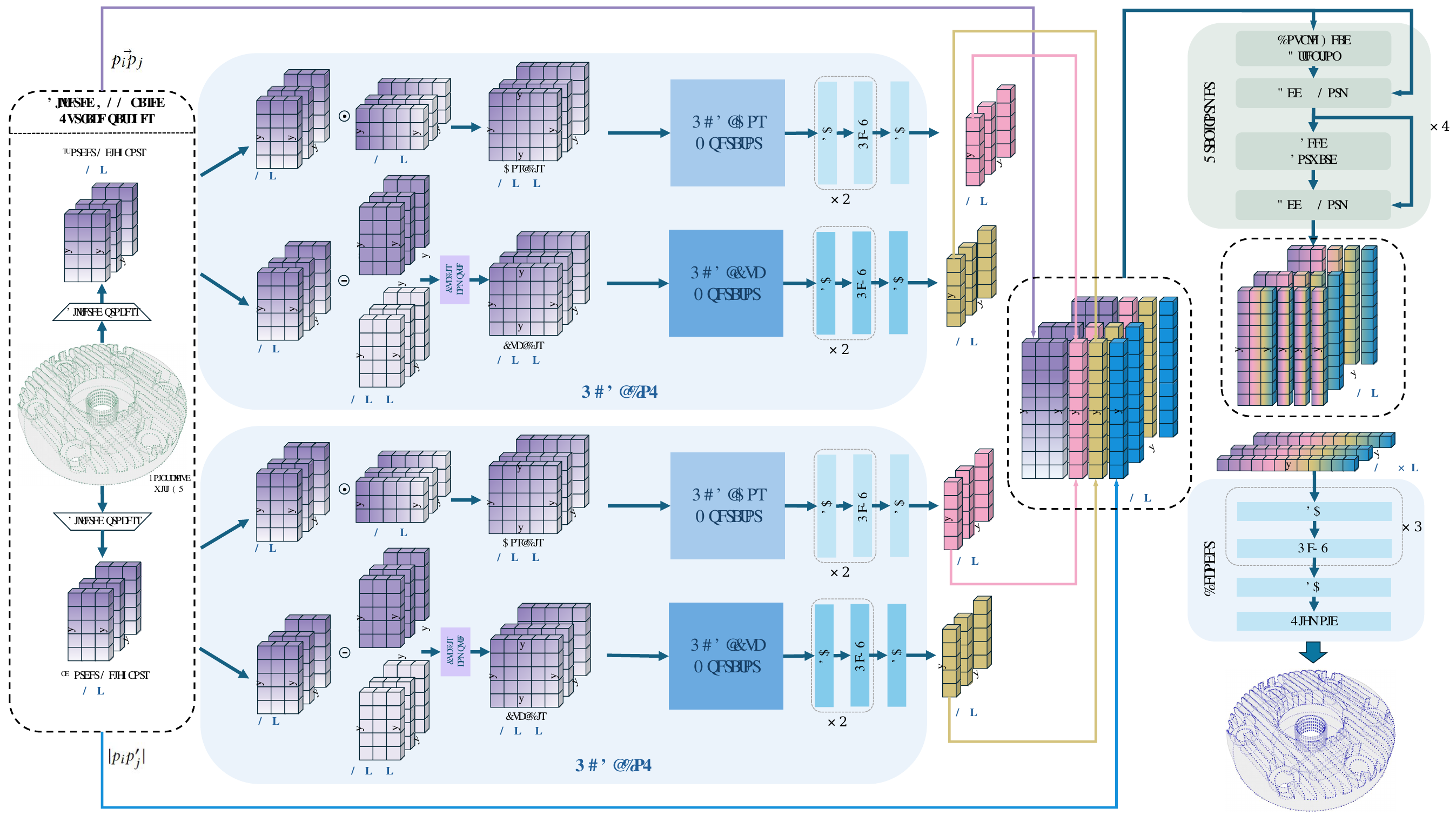}
  \caption{Overview of our proposed network OSFENet. Given a point cloud, OSFENet starts with a filtered $k$NN module for generating surface patches around each point. Then the RBF\_Dos module is  employed to process these surface patches, extracting edges $f_{EucDis}$ and $f_{CosDis}$ for each point. Subsequently, a feature map is constructed by concatenating $\vec{p_ip_j}$, $|p_ip_j'|$, $f_{EucDis}$, and $f_{CosDis}$, which is fed into Transformer encoders. Finally, the features output by the Transformer encoders are flattened and passed through a decoder to predict the probability of each point.}
  \label{fig:network}
\end{figure*}

\section{RBF-based Surface Description}
\label{DoDintroduction}

% \subsection{Preliminaries}

Radial basis function (RBF) shows significant utility in approximating multivariate functions based on scattered data, which has been proven to be quite useful in many tasks like surface reconstruction\cite{RBFSurfaceReconstruction}.

% In surface reconstruction, given a point cloud ${P} = \{p_i \in \textbf{R}^3, i= 1, 2, ..., N \}$ sampled from an unknown surface $\Gamma$, \textit{i.e.,} $P \subset \Gamma$, the goal is to find a function $f$ which implicitly defines a surface $\Gamma'$ that is a reasonable approximation to $\Gamma$. Therefore, if a surface $\Gamma$ consists of all points $p_i(x_i,y_i,z_i)$ that satisfy the Equation \eqref{equ:f=0}:

% \begin{equation}
% \label{equ:f=0}
%     f(x_i,y_i,z_i) = 0
% \end{equation}

% \noindent $f$ is considered to implicitly define $\Gamma$. 

Describing surfaces implicitly with various functions is a well-known technique. Among these approaches, RBF-based method is one of the most commonly used due to its strong fitting ability. In particular, we require that:

\begin{equation}
\label{equ:RBF}
    f(p_i)=\sum_{j=1}^N\lambda_j \phi(\left\|p_i-p_j\right\|)
\end{equation}

\noindent for all points $p_i (p_j)$ on an unknown surface $\Gamma$, where $\phi(\cdot)$ is the RBF, $||\cdot||$ is generally the Euclidean distance between points $p_i$ and $p_j$, and $\lambda_j$ are the coefficients. 

% Furthermore, several works also aimed to approximate surface derivatives via RBF\cite{RBFSurfaceGre, RBFSurfaceDerivatives}, for instance, the surface gradient. Specifically, analytically applying the differential operator $L$ to the radial function gives: 

% \begin{equation}
% \label{equ:g_L}
%     g(p_i)=\sum_{j=1}^N\lambda_j L\phi(\left\|\vec{x}_i-\vec{x}_j\right\|)
% \end{equation}

% \noindent where $g(p_i)$ is the value of the underlying function’s derivative at each $p_i$. In Equation \eqref{equ:g_L}, $\lambda_jL$ can be regarded as the expansion coefficients facilitating the computation of $g(p_i)$ through the application of $\lambda_jL$ to $\phi(||\vec{p}_i-\vec{p}_j||)$, leading to the matrix equation:

% \begin{equation}
%     \label{equ:matrix_gL}
%     \vec{\lambda'}\phi(\left\|\vec{x}_i-\vec{x}_j\right\|)=g
% \end{equation}

% \noindent where $\vec{\lambda'}$ is the vector containing $\lambda_jL$.

% \subsection{RBF-based surface description in deep learning network}

Inspired by these approaches\cite{RBFSurfaceGre, RBFSurfaceDerivatives}, we use $f(p_i)$ that is the underlying RBF function at each point $p_i$ to form the descriptor of the surface patches. The value of $\vec{\lambda}=\{\lambda_j, 0\leq \lambda_j \leq k \}$ is considered the parameters of the fully connected ($FC$) layers in the deep network, and the Equation \eqref{equ:RBF} can be formulated as:

\begin{equation}
\label{equ:MLP(f)}
    g\_{EucDis} = FC(\textbf{M}_{EucDis})
\end{equation}

\begin{equation}
\label{equ:MLP(fEuc)}
    \textbf{M}_{EucDis} = \left[\begin{array}{ccc}\phi_{1,1} & \dots & \phi_{1,N}\\ \vdots & \ddots & \vdots\\ \phi_{N,1} & \dots & \phi_{N,N}\end{array}\right].
\end{equation}

\noindent where $\phi_{i,j}=\phi(p_{i}-p_{j}\|)$ and $\phi(\cdot)$ is the Gaussian function, in this study. 

It should be noted that the RBF-based surface descriptors can be effectively learned using a single FC layer, depending on its size, as opposed to the hand-crafted descriptors employed in 3DSmoothNet \cite{3DSmoothNet} and CGF \cite{CGF}.

Euclidean distance is typically used in radial basis functions, however, alternative distance metrics are also applicable. The cosine of the angle between vectors $\vec{op_i}$ is another important distance metric between points $p_i$, which can also be used for building surface descriptors. Therefore, in this study, the cosine distances (Equation\eqref{equ:cosDis}) also serve as distance metrics in RBF for forming the surface descriptor:

\begin{equation}
\label{equ:MLP(fcos)}
    g\_{CosDis} = FC(\textbf{M}_{CosDis})
\end{equation}

\begin{equation}
\label{equ:cosDis}
    \textbf{M}_{CosDis} = \left[\begin{array}{ccc}\gamma_{1,1} & \dots & \gamma_{1,N}\\ \vdots & \ddots & \vdots\\ \gamma_{N,1} & \dots & \gamma_{N,N} \end{array}\right].
\end{equation}

\noindent where $\gamma_{i,j}=\gamma_{(\vec{op_i}\cdot \vec{op_j})}$, $\vec{op_i}$ is normalized vector and $\gamma(\cdot)$ is the cubic function, in this study.

\section{The OSFENet}
\label{Network}

\subsection{The Architecture of OSFENet}
The proposed OSFENet introduces key enhancements to the edge detection performance, specifically presenting the RBF-based surface description, which can be regarded as the features for classifying points. Specifically, given a point cloud ${P} = \{ p_i \in \textbf{R}^3, i= 1, 2, ... , N \}$ consisting of $N$ points, we generate $N$ surface patches composed by each point $p_i$ with its filtered-$k$NN $N_{k}^{p_i}$ (as detailed in Section \ref{data Preparation}). The $N$ surface patches $S=\{\vec{p_ip_j}, p_j \in N_{k}^{p_i}\}$ are taken as the input to the proposed OSFENet and the target point $p_i$ will be classified into edge point or non-edge point. The architecture of the proposed OSFENet is depicted in Figure \ref{fig:network}. 

As illustrated in Figure \ref{fig:network}, OSFENet consists of the RBF\_DoS module for generating RBF\_based surface descriptors and a 4-layer Transformer encoder \cite{Vaswani2017Attention} for encoding them. Specifically, by concatenating the output of the RBF\_DoS module with the surface patch $S$, the network incorporates information on surface descriptors, point distribution, and sampling density, which is then fed into the Transformer encoders. The decoder, comprising an MLP, ultimately outputs a probability $e_i$ for each point $p_i$ being an edge point. The following sections provide a detailed description of these modules and their implementation.

To train our network, we choose the Binary Cross Entropy (BCE) loss for predicting edge points and non-edge points, which is shown in Equation \eqref{loss}:
\begin{equation}
L_{edge} = \frac{1}{bz} \sum_{i = 1}^{bz} BCE(e_i, e_{gt})
\label{loss}
\end{equation}

\noindent where $e_i$ and $e_{gt}$ are the predicted probabilities for being an edge point and ground-truth labels of the same point, and $bz$ refers to the batch size.

\subsection{Filtered-\textit{k}NN based Local Surface Patch Representation}
\label{data Preparation}

The surface patch generation method in EC-Net \cite{Yu2018ECNet}, which creates patches containing 1024 points approximately within a geodesic radius from the centroid point, is less computationally efficient, because the method typically requires constructing a weighted graph and computing the shortest path. 

In this study, we propose a filtered $k$NN-based local surface patch representation, which can significantly reduce the number of labeled models needed for network training. This representation remains consistent for both network training and inference. 

Since the $k$NN is computed based on Euclidean distance, it may include points from opposite sides of a thin surface. To address this, we filtered out such points, resulting in a refined $k$NN, which we refer to as the filtered-$k$NN. Specifically, for each point $p_i$ on the given point cloud ${P}$, we first compute its $2k$ nearest neighboring points $N_{2k}=\{p_{j},1 \le j \le 2k\}$ based on Euclidean distance. Subsequently, we use PCA to compute the eigenvector $\vec{v}$ of $N_{2k}$ with the smallest eigenvalue. Then each point $p_{j}$ is projected on $\vec{v}$, generating the projected point ${p'_j}$. Finally, the $k$ nearest neighboring points, determined by the $k$ smallest Euclidean distances between $p_i$ and $p'_j$, form the filtered $k$NN set, denoted $N_{k}^{p_i}$. Consequently, the surface patch $S_i =\{\vec{p_ip_j}, p_j\in N_{k}^{p_i}\}$ around the target point $p_i$, is taken as the input to the network.

\subsection{Structure of the RBF\_DoS module}
\label{DoDModule}

The primary role of the RBF\_DoS module is to generate the descriptor for the surface patch $S_i$ centered around point $p_i$. It's worth noticing that the points $p_j$ within $S_i$ are categorized into two groups: 1st-order neighboring points and 2nd-order neighboring points, with $k/2$ points in each group. The points in the 1st-order group and 2nd-order group represent different scales of the neighborhood of point $p_i$, forming different shapes of the underlying surface. Subsequently, each group passes its respective RBF\_DoS module.

As outlined in Section \ref{DoDintroduction}, the first layer functions as Equations \eqref{equ:MLP(f)} and \eqref{equ:MLP(fcos)}, illustrated in Figure \ref{fig:network}. In particular, within each RBF\_DoS module, cosine distances between $\vec{p_ip_j}$, along with Euclidean distances between points $p_j$, are initially computed, forming the distance matrix $Cos\_Dis$ and $Euc\_Dis$, respectively. To facilitate distance computation, normalization of all $\vec{p_i p_j}$ is conducted, with the exclusion of the target point $p_i$ due to its coordinates being centered at 0. Following this,  RBF\_Cos and RBF\_Eus Operator are employed. One operator includes the cubic function and a $FC$ layer to process the  $Cos\_Dis$. The other operator includes the Gaussian function and a $FC$ layer to process the $Euc\_Dis$.

% Following this, two separate $FC$ layers are employed: one for the Euclidean distance matrix and another for the cosine distance matrix. 

Subsequently, two $FC$ layers with ReLU activation function are performed to obtain the surface descriptor by computing a non-linear combination of the $g\_{EucDis}$ and $g\_{CosDis}$. Finally, a $FC$ layer is utilized, forming a single value representing the feature $f_{EucDis}$ (or $f_{CosDis}$) of each point $p_j$.

Consequently, for $k$ points $p_j$ on the surface patch $S_i$, a feature of shape (\textit{k}, 6) is achieved by concatenating the $\vec{p_ip_j}$, $|p_ip_j'|$, $f_{EucDis}$ and $f_{CosDis}$. This feature map is then fed into the Transformer encoders.

\subsection{The Transformer Module}

% Generally, the determination of whether a point is an edge point or not relies on the shape of its neighborhood. 

The RBF\_DoS module generates a feature of shape ($k$,6), containing features of the $k$ nearest neighbors. Therefore, we utilize a 4-layer Transformer encoder to further encode the feature of the $k$ nearest neighboring points. It is noted that we omit positional embedding since the feature already contains information about the coordinates of points, and all points are arranged in an increasing order of $||p_i-p_j||_2$. As we use the vanilla version of the Transformer encoders, readers are referred to \cite{Vaswani2017Attention}  for more details. 

\subsection{The Decoder Layer}
The feature of shape $(k, 6)$ output by Transformer encoders is finally flattened, forming a feature of shape (1, 6*\textit{k}), which is the input to the decoder for predicting the label of the target point. As shown in Figure \ref{fig:network}, the decoder is comprised of three $FC$ layers with ReLU activation and one $FC$ layer with sigmoid activation, which outputs the probability $e_i$ for point $p_i$ being an edge point. If $e_i > 0.5$, point $p_i$ is considered an edge point, otherwise, point $p_i$ is a non-edge point.

\subsection{Network Settings}
Our network is implemented using Pytorch 1.12.1 for feature extraction as well as the neural network and its training. For computing $k$NN, the $k$NN implemented in SciPy is employed. The network is trained using the Adam optimizer with a learning rate of 1e-5. Batch size is set to 256. In each RBF\_DoS module, two $FCs$ with a structure of ($k/2$, 32) are utilized within RBF\_Cos and RBF\_Euc Operator. Then, to obtain the final surface descriptor, two $MLPs$ (16, 8, 1) are used: one generates $f_{CosDis}$ and the other generates $f_{EucDis}$. For the decoder, we adopt $MLP$ (256, 64, 32, 1).

%: One for processing the cosine distance-based RBF\_DoS, the other for processing the Euclidean similarity-based RBF\_DoS.% 
\section{Experimental Results}
\label{experiments}
In this section, we comprehensively evaluate the performance of our proposed method both quantitatively and qualitatively across a range of datasets, from CAD object models to large-scale outdoor scenes. We also compare its performance with several baselines, considering visual results, metric scores, efficiency, and parameter usage. To assess robustness, we introduce Gaussian noise and vary the sampling density of the point cloud. Additionally, results from different training models are provided to demonstrate the stability of our network. Finally, we conduct ablation studies to validate the effectiveness of the RBF\_DoS module and analyze the impact of specific design choices.

\subsection{Datasets}

Our experiments utilize datasets consisting of object-level models (such as ABC dataset\cite{Koch2019ABC}, SHREC dataset\cite{Thompson2019FeatureCE} ) and real-scanned scenes, including S3DIS\cite{Armeni20163DSP}, Semantic3D\cite{Hackel2017Semantic3DnetAN}, and UrbanBIS datasets\cite{Yang2023UrbanBISAL}. To fully utilize the training data, we augment the training data by rotating it 6 times along the X, Y, and Z axes, counterclockwise and clockwise by 90 degrees each time.

\subsubsection{Object-level Dataset}
ABC \cite{Koch2019ABC} and the SHREC dataset\cite{Thompson2019FeatureCE} both provide ground truth edges. OSFENet learns the data distribution from one model and is tested on the remaining models. The ABC dataset is utilized for performance comparison with baselines, while the SHREC dataset is employed to demonstrate limitations.

\textbf{ABC dataset}: We utilize chunk 0000 of the ABC dataset, filtering out models with missing vertex and edge information, resulting in 7071 models (denoted ABC-ALL) for testing. Furthermore, to compare with NEF\cite{NEF2023CVPR} and DEF\cite{Matveev2022DEF}, we also test OSFENet on the testing dataset in NEF \cite{NEF2023CVPR} (with 115 distinct models, denoted ABC-NEF) since no pre-trained NEF is provided and the official code of DEF\cite{Matveev2022DEF} is incomplete. 

%The ABC dataset\cite{Koch2019ABC} provides one million 3D CAD models with feature annotations.

\textbf{SHREC dataset}: The SHREC dataset\cite{Thompson2019FeatureCE} comprises 15 surfaces with annotated feature curves. Some models are obtained by scanning, some are made with $silicon$ (an online simulation platform), and some models originate from the Visionair shape workbench\cite{vvs} and the Turbosquid repository of 3D models \cite{turbosquid}. The ground truth for edges is derived from individual annotations by people from the IMATI-CNR (Italy) staff.

%  {(\url{http://visionair.ge.imati.cnr.it/})}
%  {(\url{https://www.turbosquid.com/Search/3D-Models})}

\subsubsection{Real-scanned Scene Dataset}
Sensor errors, environmental conditions, reflections, and occlusions can introduce noise and outliers, resulting in scanned data that is noisy, incomplete, or inaccurate. Moreover, real-scanned scene datasets often display significantly non-uniform point distributions. For instance, as shown in Figure \ref{teaserfigure} (b), the average sampling densities for the boundary ground, roof, and wall areas of the church are 0.035, 0.108, and 0.420, respectively. For each real-scanned scene dataset (where the ground truth feature is absent), we annotate a segment (as shown in Figure \ref{fig:realscanned}) of the scene for training the network, and the remaining scenes are then set aside for qualitative analysis evaluation.

% Furthermore, the paper of Semantic3D\cite{Hackel2017Semantic3DnetAN} also mentioned that the high measurement resolution and long measurement range lead to extreme density changes and large occlusions.

\textbf{S3DIS dataset} \cite{Armeni20163DSP}, comprises 6 indoor areas across three distinct buildings. These areas comprise diverse scenes, including bathrooms, offices, auditoriums and lounges, \textit{etc}. 

\textbf{Semantic3D dataset}\cite{Hackel2017Semantic3DnetAN} comprises real outdoor scans totaling over 4 billion points. It encompasses a variety of urban and rural scenes, including farms, town halls, sports fields, a castle, and market squares. Notably, the distribution of points within a scene is uneven, posing a challenging problem for edge extraction. 

\textbf{UrbanBIS dataset}\cite{Yang2023UrbanBISAL} consists of 6 large-scale real urban scenes, comprising a total of 2.5 billion points. These scenes spin an extensive area of 10.78 square kilometers and encompass 3,370 buildings. 

\subsection{Competitors}
We compare the performance of our OSFENet with typical baselines: BE\cite{Rusu20113D}, SGLBP\cite{Guo2022SGLBP}, EC-Net\cite{Yu2018ECNet}, PIE-NET\cite{Wang2020PieNet}, DEF\cite{Matveev2022DEF}, NerVE\cite{Nerve2023CVPR}, and  NEF\cite{NEF2023CVPR}. Since the official code of DEF\cite{Matveev2022DEF} is not available and NEF\cite{NEF2023CVPR} is a self-supervised method needing lots of time to refine, we directly use the comparison results reported in NEF\cite{NEF2023CVPR}. Furthermore, DEF\cite{Matveev2022DEF} and NEF\cite{NEF2023CVPR} will not be included in the qualitative comparison. For other methods, we extract edges based on the implementation provided by their authors and then calculate their respective metrics.

\begin{table*}
  \caption{The results on ABC-NEF \cite{NEF2023CVPR} and ABC-ALL\cite{Koch2019ABC}. The best scores are in bold.}
  \label{tab:ABCModel}
  \small
  \begin{tabular}{llccccc}
    \toprule
    Datasets & Methods & CD$\downarrow$  & IoU$\uparrow$  & Precision$\uparrow$  & Recall$\uparrow$  & F-score$\uparrow$ \\
    \midrule
    
    \multirow{8}{*}{ABC-NEF} & BE\cite{Rusu20113D} & 
    0.1319(\textcolor{mygreen}{+0.1282})& 0.1861(\textcolor{mygreen}{-78.21\%})& 0.4165(\textcolor{mygreen}{-56.42\%})& 0.6386(\textcolor{mygreen}{-34.98\%})& 0.4006(\textcolor{mygreen}{-58.69\%})\\
    
    & SGLBP\cite{Guo2022SGLBP} &  0.0344(\textcolor{mygreen}{+0.0307})& 0.8222(\textcolor{mygreen}{-14.60\%})& 0.9107(\textcolor{mygreen}{-7.00\%})& 0.9090(\textcolor{mygreen}{-7.94\%})& 0.8840(\textcolor{mygreen}{-10.35\%})\\
    
    & EC-Net\cite{Yu2018ECNet} & 0.0140(\textcolor{mygreen}{+0.0103})& 0.8833(\textcolor{mygreen}{-8.49\%})& 0.9879(\textcolor{coral}{+0.72\%})& 0.9307(\textcolor{mygreen}{-5.77\%})& 0.9578(\textcolor{mygreen}{-2.97\%})\\
    
    & PIENet\cite{Wang2020PieNet} & 0.0708(\textcolor{mygreen}{+0.0671})& 0.6709(\textcolor{mygreen}{-29.73\%})& 0.9072(\textcolor{mygreen}{-7.35\%})& 0.7204(\textcolor{mygreen}{-26.80\%})& 0.7846(\textcolor{mygreen}{-20.29\%})\\
    
    & NerVE\cite{Nerve2023CVPR} & 0.0212(\textcolor{mygreen}{+0.0175})& 0.9109(\textcolor{mygreen}{-5.73\%})&
    \textbf{0.9959}(\textcolor{coral}{+1.52\%})& 0.9284(\textcolor{mygreen}{-6.00\%})& 0.9568(\textcolor{mygreen}{-3.07\%})\\
    
    & DEF\cite{Matveev2022DEF} & 0.0402(\textcolor{mygreen}{+0.0365})& 0.7368(\textcolor{mygreen}{-23.14\%})& 0.8343(\textcolor{mygreen}{-14.64\%})& 0.7802(\textcolor{mygreen}{-20.82\%})& 0.8009(\textcolor{mygreen}{-18.66\%})\\
    
    & NEF \cite{NEF2023CVPR} & 
    0.0353(\textcolor{mygreen}{+0.0316})& 0.8283(\textcolor{mygreen}{-13.99\%})& 0.9387(\textcolor{mygreen}{-4.20\%})& 0.8838(\textcolor{mygreen}{-10.46\%})& 0.9044(\textcolor{mygreen}{-8.31\%})\\
    
    & OSFENet(Ours) & \textbf{0.0037} & \textbf{0.9682} & 0.9807 & \textbf{0.9884} & \textbf{0.9875}\\
    \midrule

    \multirow{6}{*}{ABC-ALL} & BE\cite{Rusu20113D} & 0.1233(\textcolor{mygreen}{+0.0988}) & 0.2797(\textcolor{mygreen}{-55.57\%}) & 0.5116(\textcolor{mygreen}{-38.39\%}) & 0.6087(\textcolor{mygreen}{-32.95\%}) & 0.4685(\textcolor{mygreen}{-43.12\%})\\
    
    & SGLBP\cite{Guo2022SGLBP} & 
    0.0558(\textcolor{mygreen}{+0.0313}) & 0.6088(\textcolor{mygreen}{-22.66\%}) & 0.7070(\textcolor{mygreen}{-18.85\%}) & 
    0.9097(\textcolor{mygreen}{-2.85\%}) &
    0.7442(\textcolor{mygreen}{-15.55\%})\\
    
    & EC-Net\cite{Yu2018ECNet} & 
    0.0436(\textcolor{mygreen}{+0.0191}) & 0.6295(\textcolor{mygreen}{-20.59\%}) & 0.8884(\textcolor{mygreen}{-0.71\%}) & 0.8129(\textcolor{mygreen}{-12.53\%}) & 0.8245(\textcolor{mygreen}{-7.52\%})\\
    
    & PIENet\cite{Wang2020PieNet} & 
    0.1760(\textcolor{mygreen}{+0.1515}) & 0.0887(\textcolor{mygreen}{-74.67\%}) & 0.3110(\textcolor{mygreen}{-58.45\%}) & 0.1869(\textcolor{mygreen}{-75.13\%}) & 0.1892(\textcolor{mygreen}{-71.05\%})\\
    
    & NerVE\cite{Nerve2023CVPR} & 
    0.0409(\textcolor{mygreen}{+0.0164}) & 0.7263(\textcolor{mygreen}{-10.91\%}) & 0.8813(\textcolor{mygreen}{-1.42\%}) & 0.8235(\textcolor{mygreen}{-11.47\%}) & 0.8409(\textcolor{mygreen}{-5.88\%})\\
    
    & OSFENet(Ours) & \textbf{0.0245} & \textbf{0.8354} & \textbf{0.8955} & \textbf{0.9382} & \textbf{0.8997}\\
    
    \bottomrule

  \end{tabular}
\end{table*}

\subsection{Evaluation Metrics}
\label{sec:metrics}
We use five commonly used metrics, including CD (Chamfer Distance), the IoU (Intersection over Union), Precision, Recall, and F-score, to measure the prediction of our network and that of the selected baselines. For comparison, we normalize all predicted feature points and ground truth into the range of [0, 1] and finally align them. Therefore, when computing IoU, precision, recall, and F-score, points are considered matched if there is at least one ground truth point with an L2 distance smaller than 0.02.

\textbf{CD}  (Chamfer distance) is a metric used to assess the similarity between two point sets, which is defined as:

\begin{equation*}
    CD_{(S_1,S_2)}=\frac{1}{|S_1|}\sum_{x\in S_1}\min_{y\in S_2}\|x-y\|+\frac{1}{|S_2|}\sum_{y\in S_2}\min_{x\in S_1}\|y-x\|.
\end{equation*}

For two given point sets, $S_1$ and $S_1$, the Chamfer distance is computed as the average sum of the distances from each point in $S_1$ to its closest neighbor in $S_2$, combined with the average sum of the distances from each point in $S_2$ to its nearest neighbor in $S_2$.

\textbf{IoU} (Intersection over Union, also known as Jaccard index) quantifies the degree of overlap between the predicted set and a reference set, yielding a value between 0 and 1. An IoU score of 1 indicates that the predicted set and the reference set perfectly overlap, meaning their union is identical to their intersection. Conversely, an IoU score of 0 signifies no overlap between the predicted and reference sets. The IoU score is mathematically defined as follows:

\begin{equation*}
    IoU = \frac{TP}{TP + FP + FN}.
\end{equation*}

\textbf{Precision}, also known as positive predictive value (PPV), assesses the proportion of positive identifications that are accurate. It is defined as:

\begin{equation*}
    Precision = \frac{TP}{TP + FP}.
\end{equation*}

\textbf{Recall} also referred to as true positive rate (TPR), quantifies the proportion of actual positive instances that are correctly identified by the model. A higher recall value signifies improved performance in capturing positive cases. Mathematically, recall is defined as:

\begin{equation*}
    Recall = \frac{TP}{TP + FN}.
\end{equation*}

\textbf{F-score} is a metric that combines precision and recall by calculating their harmonic mean. It is defined as:

\begin{equation*}
    F\text{-}score = 2 \times \frac{precision \times recall}{precision + recall}.
\end{equation*}

\begin{figure*}
  \centering
  \includegraphics[width=0.95\linewidth]{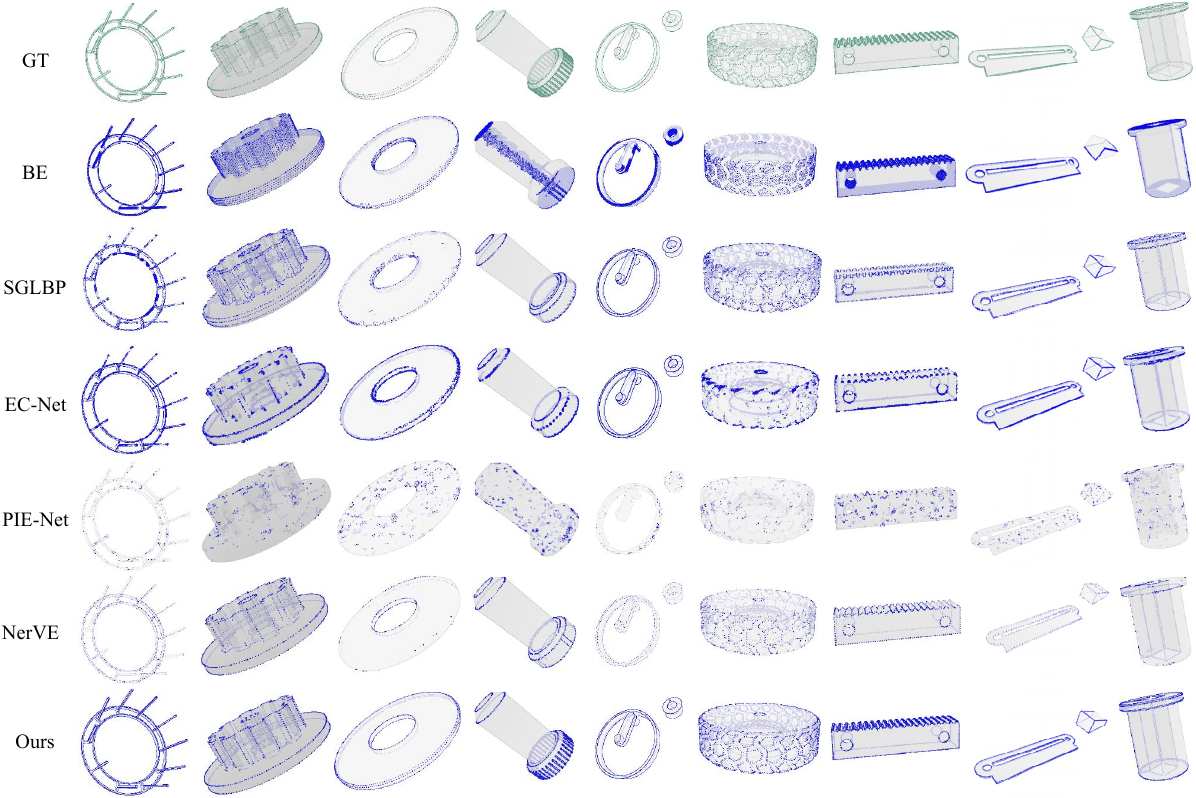}
  \caption{Qualitative comparisons on the feature points prediction with selected baselines.}
  \label{fig:qualitative comparison}
\end{figure*}

\begin{figure*}
  \centering
  \includegraphics[width=\linewidth]{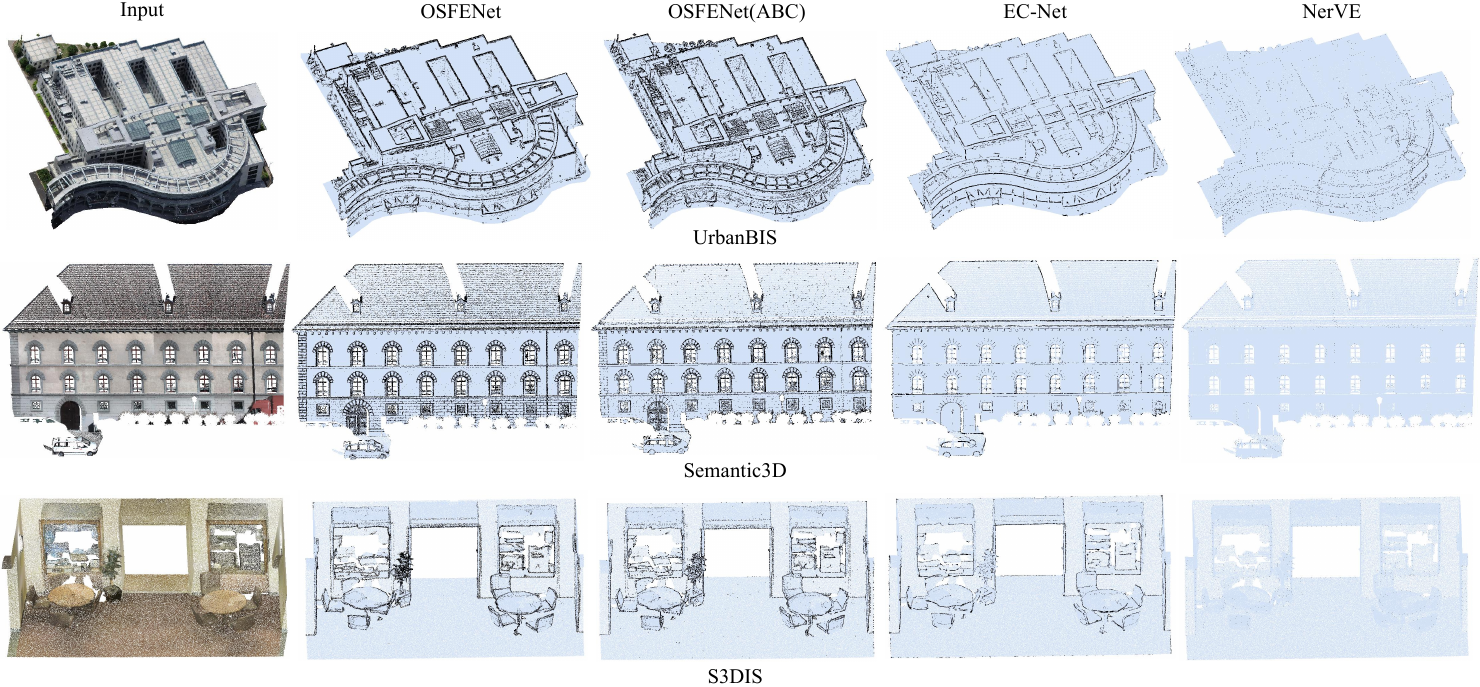}
  \caption{Comparisons are made on the real-scanned datasets generated by OSFENet (ours), EC-Net \cite{Yu2018ECNet}, and NerVE \cite{Nerve2023CVPR}. OSFENet (ABC) refers to the results obtained from training OSFENet on the ABC dataset, while OSFENet refers to the results of training on the corresponding real-scanned dataset.}
  \label{fig:comparison_realscan}
\end{figure*}

\subsection{Comparison with the State of the Art}

\subsubsection{Quantitative evaluation}

Table \ref{tab:ABCModel} lists scores of the 5 metrics on the ABC-NEF dataset (115 models) and ABC-All dataset (7071 models), respectively. As can be seen from Table \ref{tab:ABCModel}, when dealing with ABC-NEF, our approach performs better than all baselines in all metrics except for the precision of NerVE\cite{Nerve2023CVPR} and EC-Net\cite{Yu2018ECNet}. The ABC-NEF dataset contains 85 models (of 115 test models) that are training models of NerVE\cite{Nerve2023CVPR}; nevertheless, our approach still outperforms NerVE and EC-Net in CD, IoU, Recall, and F-score. Furthermore, when dealing with ABC-ALL, scores also demonstrate that the proposed OSFENet outperforms all the baselines across the five metrics.

\subsubsection{Qualitative comparison}

Figure \ref{fig:qualitative comparison} displays the edge points detected by the baselines, excluding DEF\cite{Matveev2022DEF} and NEF\cite{NEF2023CVPR}, alongside ours. It can be found that for methods such as BE\cite{Rusu20113D}, SGLBP\cite{Guo2022SGLBP}, EC-Net\cite{Yu2018ECNet}, PIENet\cite{Wang2020PieNet}, many non-edge points in the vicinity of edges are erroneously taken as edge points, especially in the case of some small and complex 3D point clouds. Moreover, in the case of point clouds with massive fine edges like table legs, these methods fail to distinguish edges from non-edges. For the most recently proposed approach NerVE\cite{Nerve2023CVPR}, finer structured edges are inevitably lost due to the volumetric grid representation used. However, our method is able to identify edge points on point clouds with complex structures, diverse shapes, and finer-grained details.

\subsubsection{Comparison on real-scanned datasets}

We choose EC-Net\cite{Yu2018ECNet} and NerVE\cite{Nerve2023CVPR} as baselines for comparison on real-scanned datasets, as they achieved the first and second highest scores on the ABC dataset. To ensure a fair comparison, we test our OSFENet by training it on the ABC dataset and the corresponding scene dataset, respectively, and then evaluating it on the scenes. Figure \ref{fig:comparison_realscan} presents the testing results.

%of EC-Net\cite{Yu2018ECNet} and NerVE\cite{Nerve2023CVPR}, and that of our OSFENet, on the real-scanned datasets.}

It can be observed that EC-Net\cite{Yu2018ECNet} primarily extracts sharp edges while failing to capture the smooth ones, as it predominantly focuses on the former. Unlike the ABC dataset, indoor and outdoor scenes display a wider range of scales and diverse levels of detailed edges, presenting additional challenges. Regarding the most recent method, NerVE\cite{Nerve2023CVPR}, nearly all edge points are disregarded due to the normalization and voxelization operations. Achieving effective results becomes difficult when attempting to set a consistent voxel scale for models of various sizes, particularly for outdoor scenes. However, our OSFENet consistently outperforms the selected baselines, with better results when trained on the scene dataset compared to the ABC dataset. This highlights the advantage of learning the target point cloud's specific data distribution over general data distributions.

\subsubsection{Efficiency}

Table \ref{tab:efficiency} presents the execution timings and throughput of our OSFENet compared to EC-Net\cite{Yu2018ECNet} and NerVE\cite{Nerve2023CVPR}. Specifically, we evaluate the efficiency of OSFENet and the competitors across a range of model sizes, from eight thousand points to ten million points. For models with fewer than one million points, our method is faster than EC-Net\cite{Yu2018ECNet} but slower than NerVE\cite{Nerve2023CVPR}. However, when dealing with models larger than three million points, both NerVE\cite{Nerve2023CVPR} and EC-Net\cite{Yu2018ECNet} fail to provide valid results, while our method still produces results within an acceptable timeframe. Furthermore, we compute the model throughput by measuring the number of points processed per second (PPS). As shown in Table \ref{tab:efficiency}, our OSFENet outperforms the selected baselines in terms of throughput.

% Since the input to PIENet \cite{Wang2020PieNet} is down-sampled to 4096 points, PIENet is not included in the efficiency comparison. 

\begin{table}
  \caption{The execution timings (in seconds) of ours compared with two competitors. The timings are in seconds.  {The throughput calculation only takes into account the inference time of the learning model.}}
  \label{tab:efficiency}
  \begin{tabular}{lllll}
    \toprule
    Models &  \#Points & OSFENet(Ours) & NerVE & EC-Net \\
    \midrule
    Bookshelf & 8845 & 1.7849  & 0.6873  & 2.9348 \\
    Nut & 22256 & 3.1856 & 0.7390  & 6.9550  \\
    Plane & 182644 & 11.8553  & 1.3639  & 50.4284 \\
    Train & 626785 & 22.9518 &3.1400 & 136.6834 \\
    Office\_1 & 1104420 & 30.9747 & 6.2146  &  256.4788 \\
    Office\_2 & 3804929 & 79.3522 & - & -\\
    Qingdao & 12748044 & 241.7295 & - & -\\
    \midrule
    \multicolumn{2}{l}{ {Throughput (PPS)}} &  {502k} &  {230k} &  {76k} \\
    \bottomrule
  \end{tabular}
\end{table}

\begin{table}
  \caption{The scales of training datasets and parameter specifications between competitors and ours. '-' indicates the absence of values.}
  \label{tab:trainingsets}
  \small
  \begin{tabular}{lll}
    \toprule
    Methods & \#Training Model & \#Parameters (M) \\
    \midrule
    EC-Net\cite{Yu2018ECNet} & 36 & 0.82  \\
    PIENet\cite{Wang2020PieNet} & 32 & 1.03    \\
    NerVE\cite{Nerve2023CVPR} & 1892 & 4.08   \\
    DEF\cite{Matveev2022DEF} & - & 78.78   \\
    NEF \cite{NEF2023CVPR} & - & 1.58    \\
    OSFENet(Ours) & \textbf{1} & \textbf{0.04}   \\
    \bottomrule
  \end{tabular}
\end{table}

\subsubsection{Parameters}

Table \ref{tab:trainingsets} presents the scale of training models and network parameters of both our network and those of our competitors. It can be seen that PIENet\cite{Wang2020PieNet} utilizes the fewest models for training among the selected baseline methods. However, its performance is not competitive. Furthermore, NEF\cite{NEF2023CVPR} is a self-supervised model that requires parameter tuning for each model. However, despite utilizing only 1 training model for training, our OSFENet still achieves the highest scores.

For parameters, EC-Net\cite{Yu2018ECNet} occupies the fewest parameters among the baselines, followed by PIENet\cite{Wang2020PieNet}, although its scale is significantly larger than ours.

\subsection{Robustness}

\begin{table}
  \caption{Scores on noisy models and down-sampled models in ABC-ALL dataset. 'N\_' stands for noisy models and 'S\_' stands for down-sampled models. Scores higher than those of competitors are highlighted in \textcolor{coral}{coral}.}
    \label{tab:noisy}
    \small
  \begin{tabular}{lccccc}
    \toprule
    Models & CD$\downarrow$  & IoU$\uparrow$ & Precision$\uparrow$  & Recall$\uparrow$ & F-score$\uparrow$ \\
    \midrule
    N\_0.03 & \textcolor{coral}{0.0316} & \textcolor{coral}{0.7798} & 0.8637 & 0.9085 & \textcolor{coral}{0.8635}\\
    
    N\_0.05 & \textcolor{coral}{0.0382} & \textcolor{coral}{0.7504} & 0.8162 & \textcolor{coral}{0.9229} & \textcolor{coral}{0.8446} \\
    
    N\_0.10 & \textcolor{coral}{0.0397} & \textcolor{coral}{0.7377} & 0.8075 & \textcolor{coral}{0.9199} & 0.8405\\
   
    S\_0.90 & \textcolor{coral}{0.0283} & \textcolor{coral}{0.7993} & 0.8829 & \textcolor{coral}{0.9120} & \textcolor{coral}{0.8772}\\
    
    S\_0.75 & \textcolor{coral}{0.0365} 
    & \textcolor{coral}{0.7533} & 0.8074 & \textcolor{coral}{0.9347} & \textcolor{coral}{0.8434}\\
    
    S\_0.60 & \textcolor{coral}{0.0405} & 0.6706 & 0.8039 & \textcolor{coral}{0.8290} & 0.7863\\
    \bottomrule
   \end{tabular}
\end{table}

% In addition to the quantitative and qualitative evaluation, we assess the robustness of our method by introducing Gaussian noise or varying the sampling density of the input point cloud. The results are also shown in Table \ref{tab:diffModels}. For introducing Gaussian noise, the disturbance radius for generating various scales of Gaussian noise is set as $\sigma = 0.03Sd$, $0.05Sd$, $0.1Sd$, $Sd$ is the average Euclidean distances between points and their $k$NN ($k=16$) on the point cloud. As observed, even with slight noise such as $\sigma = 0.03Sd$, $0.05Sd$, our CD, IoU,  and F-score scores still outperform other baselines when dealing with clean data. 

% For downsampling the models, we use random sampling implemented in NumPy to downsample point clouds with sampling ratios of 0.9, 0.75, and 0.6, respectively. The results listed in Table \ref{tab:diffModels} indicate that our network, trained on resampled point clouds at ratios of 0.9 or 0.75, consistently outperforms by comparing with competitors, which are performed on data without noise and down-sampling.

In addition to the quantitative and qualitative evaluations, we further assess the robustness of our method by introducing Gaussian noise and varying the sampling density of the input point cloud. The results are summarized in Table \ref{tab:noisy}. 

For Gaussian noise, we perturb the point cloud by adding noise with disturbance radii of $\sigma = 0.03Sd$, $0.05Sd$, and $0.1Sd$, where $Sd$ is the average Euclidean distance between points and their $k$-nearest neighbors ($k=16$). As shown in Table \ref{tab:noisy}, even at low noise levels ($\sigma = 0.03Sd$ and $0.05Sd$), our method consistently outperforms baselines on CD, IoU, and F-score metrics, which were tested on clean data.

In the downsampling experiments, we used random sampling with ratios of 0.9, 0.75, and 0.6 to reduce point cloud density. As shown in Table \ref{tab:noisy}, our network consistently outperforms competing methods when trained on resampled point clouds at ratios of 0.9 or 0.75, demonstrating its robustness and adaptability across diverse data conditions.

\subsection{Results of different training models}
\begin{figure}
  \centering
  \includegraphics[width=\linewidth]{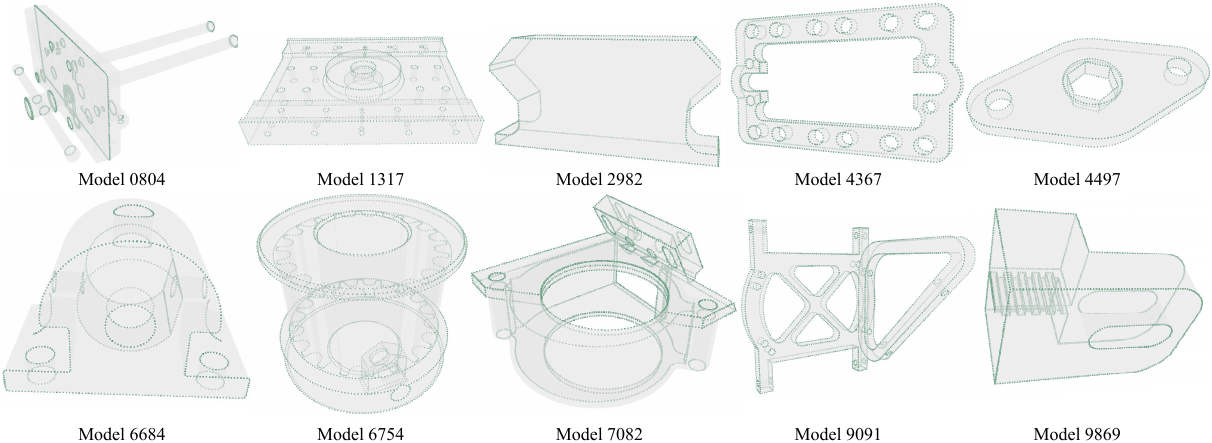}
  \caption{Different training models for the ABC dataset\cite{Koch2019ABC}.}
  \label{fig:diffModels}
\end{figure}

The key idea of this study is to learn the specific data distribution of the target point cloud, and thus achieve superior results than networks that were trained on general data distributions. Therefore, we use different target models to train the proposed OSFENet and then evaluate the performance in ABC-All dataset. The selected models are shown in Figure \ref{fig:diffModels} and the scores are listed in Table \ref{tab:diffModels}. It can be observed that when using different training models, the performance of the proposed OSFENet remains superior to that of the competitors.

\subsection{Additional results on real-scanned datasets}

% For a specific dataset, by learning the labeled segment of the scene, the proposed OSFENet can achieve satisfactory results for predicting the edges of the models from that dataset, allowing for edge extraction on datasets with a limited number of models as well as large-scale scenes (comprising more than ten million points) without significant annotation pressure. 

The proposed OSFENet can predict model edges effectively for specific datasets, enabling edge extraction on both small datasets and large-scale scenes (over ten million points) with minimal annotation effort.

\begin{table}
  \caption{Scores achieved using different training models in ABC-ALL dataset.}
\label{tab:diffModels}
  \begin{tabular}{lccccc}
    \toprule
    Training model & CD$\downarrow$ & IoU$\uparrow$ & Precision$\uparrow$ & Recall$\uparrow$ & F-score$\uparrow$\\
    \midrule
     Model 0804 & 0.0245 & 0.7757 & 0.9371 & 0.8513 & 0.8738\\
     Model 1317 & 0.0299 & 0.7648 & 0.9064 & 0.8555 & 0.8583\\
     Model 2982 & 0.0303 & 0.7735 & 0.8955 & 0.8715 & 0.8596\\
     Model 4367 & 0.0283 & 0.8053 & 0.8875 & 0.9138 & 0.8769\\
     Model 4497 & 0.0298 & 0.7947 & 0.8868 & 0.9102 & 0.8709\\
     %  {Model 6089} & 0.0382 & 0.7420 & 0.7923 & 0.9388 & 0.8246\\
     Model 6684 & 0.0300 & 0.7730 & 0.9024 & 0.8658 & 0.8616\\
     Model 6754 & 0.0282 & 0.7908 & 0.9007 & 0.8865 & 0.8726\\
     Model 7082 & 0.0254 & 0.7983 & 0.9055 & 0.8967 & 0.8845\\
     Model 9091 & 0.0265 & 0.8079 & 0.9002 & 0.9054 & 0.8829\\
     Model 9869 & 0.0265 & 0.8195 & 0.8906 & 0.9264 & 0.8875\\
     
     \bottomrule
  \end{tabular}
\end{table}

% including indoor scene dataset S3DIS\cite{Armeni20163DSP}, and outdoor scene datasets including Semantic3D\cite{Hackel2017Semantic3DnetAN}, and UrbanBIS\cite{Yang2023UrbanBISAL}

\begin{figure*}[tb]
  \centering
  \includegraphics[width=0.95\linewidth]{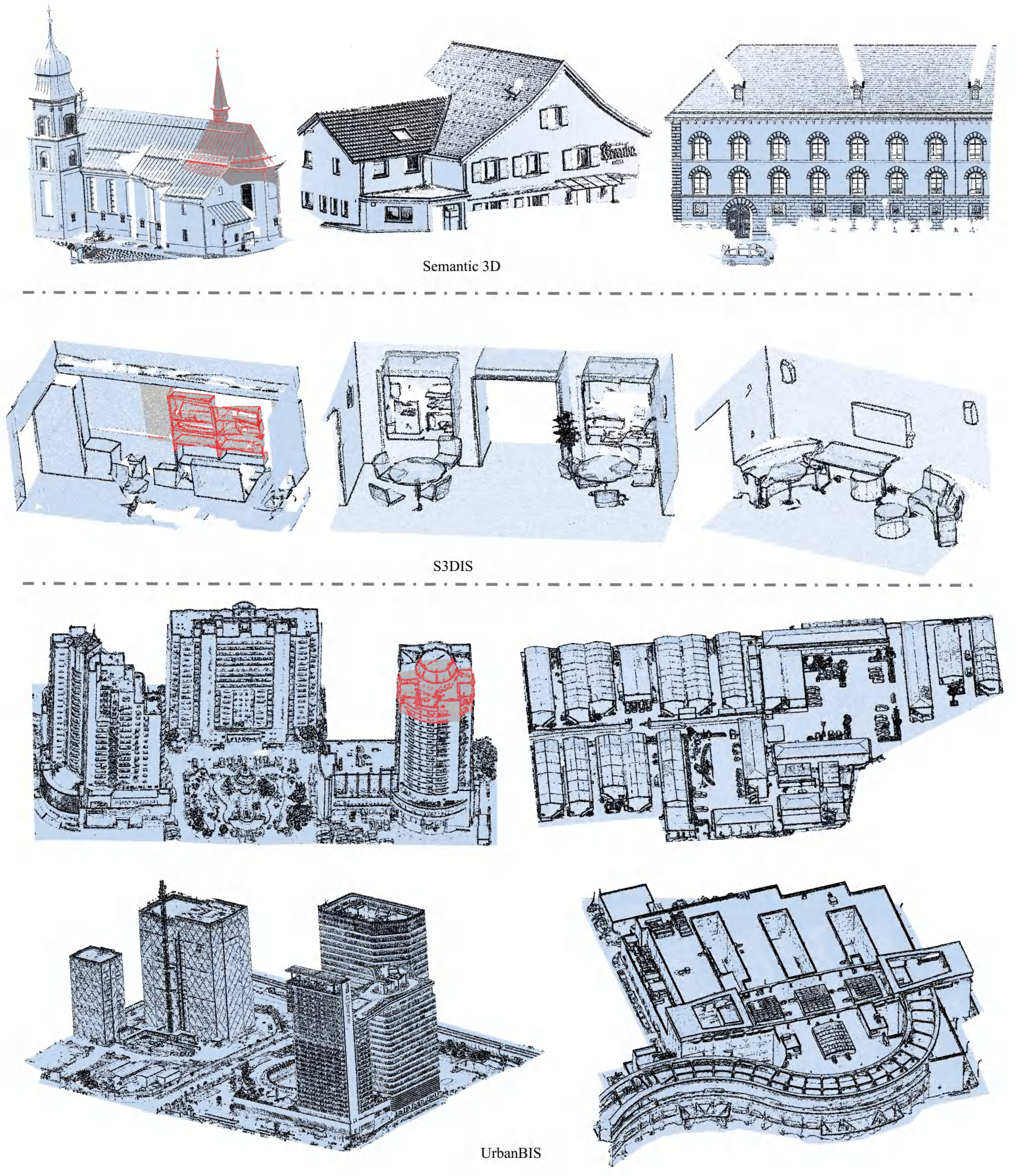}
  \caption{Results of edge detection on the real-scanned datasets. The red points and gray points are training edges and non-edges of the corresponding dataset, and the black points are the predicted edges obtained by the proposed OSFENet.} 
  \label{fig:realscanned}
\end{figure*}

\clearpage

In Figure \ref{fig:realscanned}, we showcase the results of the proposed OSFENet on several real-scanned datasets. Indoor and outdoor scenes differ in various aspects such as scale, shape, level of detail, sampling density, and noise distribution. Moreover, the buildings in Semantic3D exhibit both sharp and smooth edges, whereas the buildings in UrbanBIS showcase more massive and detailed edges. It can be seen that for each dataset, the proposed OSFENet effectively learns from the training segment respectively, and consistently produces satisfactory predictions. 

\subsection{Ablation Study}

\begin{table}
  \caption{Results of ablation studies on ABC-NEF dataset. The best scores are in bold, and scores higher than those of competitors are highlighted in \textcolor{coral}{coral}.}
  \label{tab:ablation test}
  \small
  \begin{tabular}{lccc}
    \toprule
    Methods & CD$\downarrow$  & IoU$\uparrow$  & F-score$\uparrow$ \\
    \midrule
    w/o RBF\_DoS & 0.0679 & 0.4139 & 0.5961\\
    
    w/o Transformer & 0.0182 & 0.8346 & 0.9096 \\
    \midrule    
    
    16NN+$f_{CosDis}$+$f_{EucDis}$ & 0.0283 & 0.7469 & 0.8533\\
    
    16NN+$f_{CosDis}$+$|p_ip_j'|$ & 0.0262 & 0.7668 & 0.8678\\
    
    16NN+$f_{EucDis}$+$|p_ip_j'|$ & 0.0198& 0.7639 & 0.8790\\
    $f_{CosDis}$+$f_{EucDis}$+$|p_ip_j'|$ & 0.0160 & 0.8592 & 0.9244\\
    
    16NN+$f_{CosDis}$ & 0.0429 & 0.6094 & 0.7513\\ 
    
    16NN+$f_{EucDis}$ & 0.0416 & 0.6197 & 0.7632\\
    
    16NN+$|p_ip_j'|$ & 0.0360 & 0.6564 & 0.7997\\
    \midrule
    
    8NN &\textcolor{coral}{0.0116} & 0.8929 & 0.9474\\
    
    24NN & \textcolor{coral}{0.0060} & \textcolor{coral}{0.9495} & \textcolor{coral}{0.9707}\\
    \midrule
    
    OSFENet(ours) & \textbf{0.0037} & \textbf{0.9682} & \textbf{0.9875} \\
    \bottomrule
  \end{tabular}
\end{table}

In this section, we validate the effectiveness of RBF\_DoS and our design choices by performing ablation studies on the ABC-NEF dataset, and the quantitative results are presented in Table \ref{tab:ablation test}.

The main enhancement of performance attributes to the introduction of the RBF\_DoS module, which integrates RBF-based surface descriptor with deep learning. Therefore, we first conduct ablation studies to validate the necessity of the RBF\_DoS module by removing RBF\_DoS module (w/o RBF\_DoS) and transformer module (w/o Transformer), respectively. It can be observed from Table \ref{tab:ablation test} that all scores experience a notable decrease upon the removal of RBF\_DoS. However, when the transformer is removed, the scores only marginally decrease but are still higher than some of the competitors, underscoring the effectiveness and significance of the proposed RBF\_DoS module.

To further assess the necessity of the RBF\_DoS module, we conducted comprehensive ablation studies by testing different combinations of $f_{CosDis}$, $f_{EucDis}$, $\vec{p_ip_j}$ and $|p_ip_j'|$, separately. The scores presented in Table \ref{tab:ablation test} reveal that all pieces of information are necessary and important for distinguishing edges from non-edges, with $f_{CosDis}$ and $f_{EucDis}$ playing particularly crucial roles.

Furthermore, $k$NN is also an important setting for the proposed method. In the experiments in the main paper, we set $k=16$, and here, we set $k=8$(8NN) and $k=24$(24NN) for another group of ablation studies about the required number of points forming the surface patch to train OSFENet properly. It can be observed in Table \ref{tab:ablation test} that better performance is achieved by 16NN compared to those by 8NN and 24NN. When $k=8$, the reduced number of neighboring points provides less context for the surface patch, resulting in a decline in edge detection performance. Conversely, when $k=24$, the larger scale surface patches may extend beyond the edge region, also leading to decreased performance. However, when setting $k=8$ and $k=24$,  OSFENet still matches or even outperforms the other competitors. 

% OSFENet (ours) represents the baseline of the proposed OSFENet in the main paper. The scores obtained strongly validate the significance of the proposed RBF\_DoS and affirm the validity of our design choices.

\subsection{Surface segmentation results}

\begin{figure}[H]
  \centering
  \includegraphics[width=\linewidth]{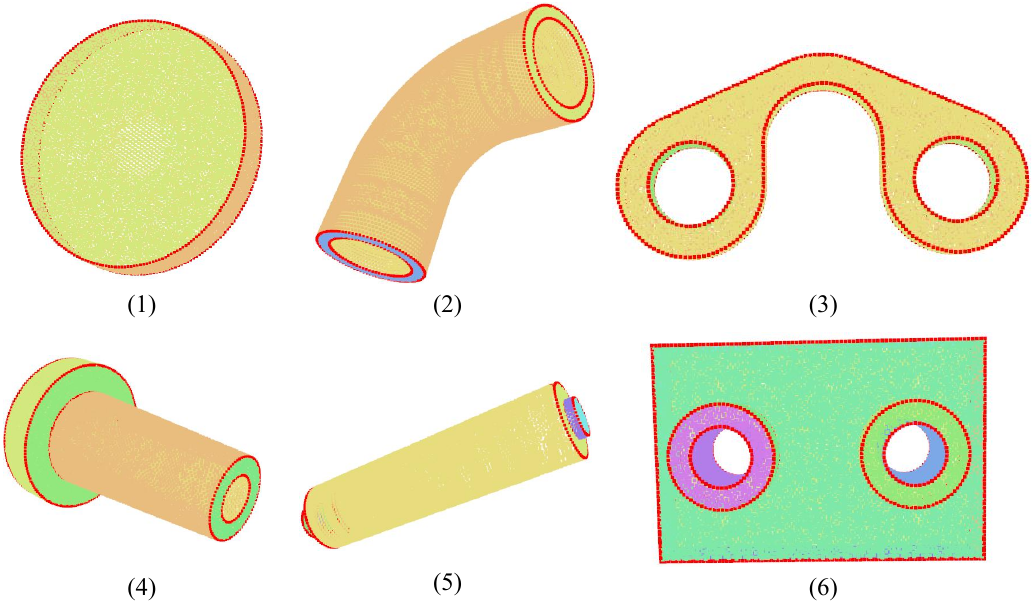}
  \caption{The segmented surfaces.}
  \label{fig:surface}
\end{figure}

Our method facilitates surface segmentation by utilizing the predicted edge points as boundaries and employing  {Breadth-First Search (BFS)} to execute the segmentation process. Specifically, the $k$-NN graph ($k$ = 5) is constructed over the point cloud, the BFS starting from a random non-visited and non-edged point (called seed point), and stopping when encountering an edge point, is performed repeatedly until no seed point can be found.  {Figure \ref{fig:surface} illustrates examples of surface segments, while the quantitative analysis scores are provided in Table \ref{tab:segresults}. The scores are computed based on the methods outlined in Section \ref{sec:metrics}. It can be observed that as the edges are accurately detected, the surface segmentation results are satisfactory, which further highlights the utility of the proposed method. However, this straightforward flood-filling approach to surface segmentation may fail in scenarios where small gaps are present in the edges (failure cases are provided in the supplementary materials).}

\begin{table}
  \caption{Surface segmentation results}
  \label{tab:segresults}
  \small
  \begin{tabular}{llllll}
    \toprule
     {Models} &  {CD$\downarrow$}  &  {IoU$\uparrow$} &  {Precision$\uparrow$}  &  {Recall$\uparrow$} &  {F-score$\uparrow$} \\
    \midrule
     { Model(1) } &  { 0 } &  { 1.00 } &  {1.00} &  {1.00} &  {1.00}  \\
     {Model(2) } &  { 0 } &  { 1.00} &  {1.00} &  {1.00} &  {1.00}  \\
     { Model(3) } &  { 9.59e-6 } &  { 0.9988} &  {1.00} &  {0.9988} &  {0.9994}  \\
     { Model(4) } &  { 1.40e-6 } &  { 0.9998} &  {0.9999} &  {0.9999} &  {0.9999}  \\
     { Model(5) } &  { 5.74e-6 } &  { 0.9993} &  {0.9998} &  {0.9995} &  {0.9997}  \\
     { Model(6) } &  { 5.48e-7 } &  { 0.9999} &  {1.00} &  {0.9999} &  {1.00}  \\
    \bottomrule
  \end{tabular}
\end{table}
\section{Discussion on Limitations}

The main idea is to learn the point distribution characteristic of a specific dataset to achieve distinguished performance compared to models trained on datasets compiled using different techniques. We have validated our method on both CAD model datasets and real-scanned datasets, and the superior results demonstrate its effectiveness and advantage. 

\begin{figure}
  \centering
  \includegraphics[width=\linewidth]{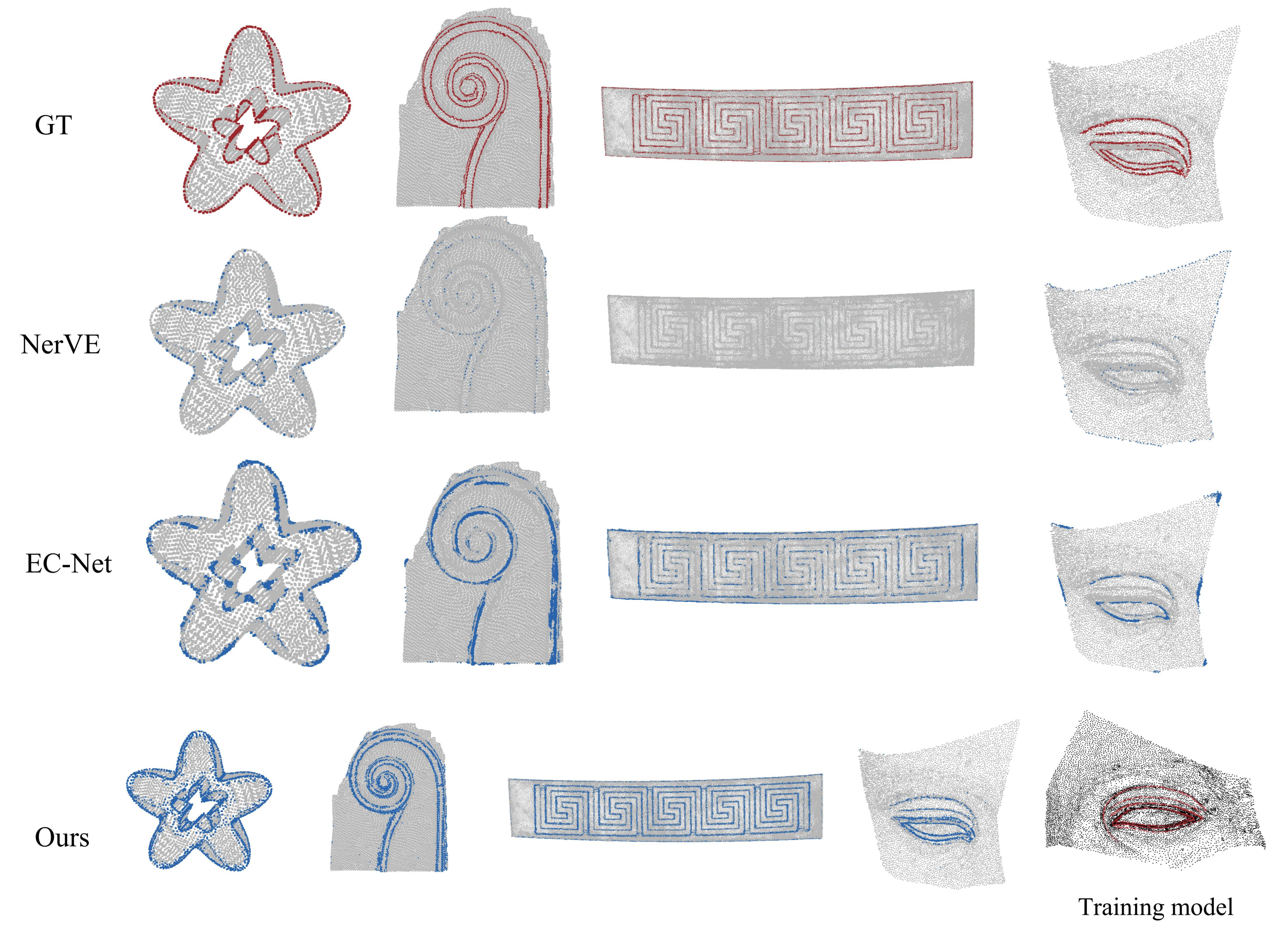}
  \caption{The results of OSFENet, EC-Net\cite{Yu2018ECNet}, and NerVE\cite{Nerve2023CVPR} on the SHREC dataset\cite{Thompson2019FeatureCE}.}
  \label{fig:limitation}
\end{figure}

However, to further validate the performance of our OSFENet, we also tested it on another object-level dataset, \textit{i.e.} SHREC, which consists of models acquired by different scanners and techniques. The training model and the predictions of our OSFENet, EC-Net\cite{Yu2018ECNet}, and NerVE\cite{Nerve2023CVPR} are presented in Figure \ref{fig:limitation}. It can be seen that although some non-edges in the edge region are mistakenly identified as edge points, our method still accurately extracts the sharp edges of the models. However, both EC-Net \cite{Yu2018ECNet} and NerVE \cite{Nerve2023CVPR} fail to extract sufficient sharp edges describing the silhouettes of the models. This supports the assertion that training a network on a dataset with data from different scanners is less effective than using data from a single scanner.

\section{Conclusions}
\label{conclusion}
In this study, we introduce OSFENet, a one-shot learning-based network designed for feature extraction on point clouds. Our primary contributions lie in the development of a one-shot learning framework for point cloud feature extraction and the introduction of the RBF\_DoS module. This framework allows for network training on datasets of limited size and large-scale scenes without significant annotation pressure. Experimental results demonstrate that our method surpasses existing approaches and exhibits prospects for handling outdoor scenes.

One limitation of our framework arises when dealing with datasets comprising models acquired via varying methods. The necessity and effectiveness of the RBF\_DoS module are validated by results from various datasets and ablation studies. Extending RBF\_DoS to encompass the entire shape of point clouds for tasks such as classification and segmentation presents an interesting and valuable point for future research.

\bibliographystyle{IEEEtran}
\footnotesize
\bibliography{ref}

\clearpage

\begin{figure*}
  \centering
  \includegraphics[width=\linewidth]{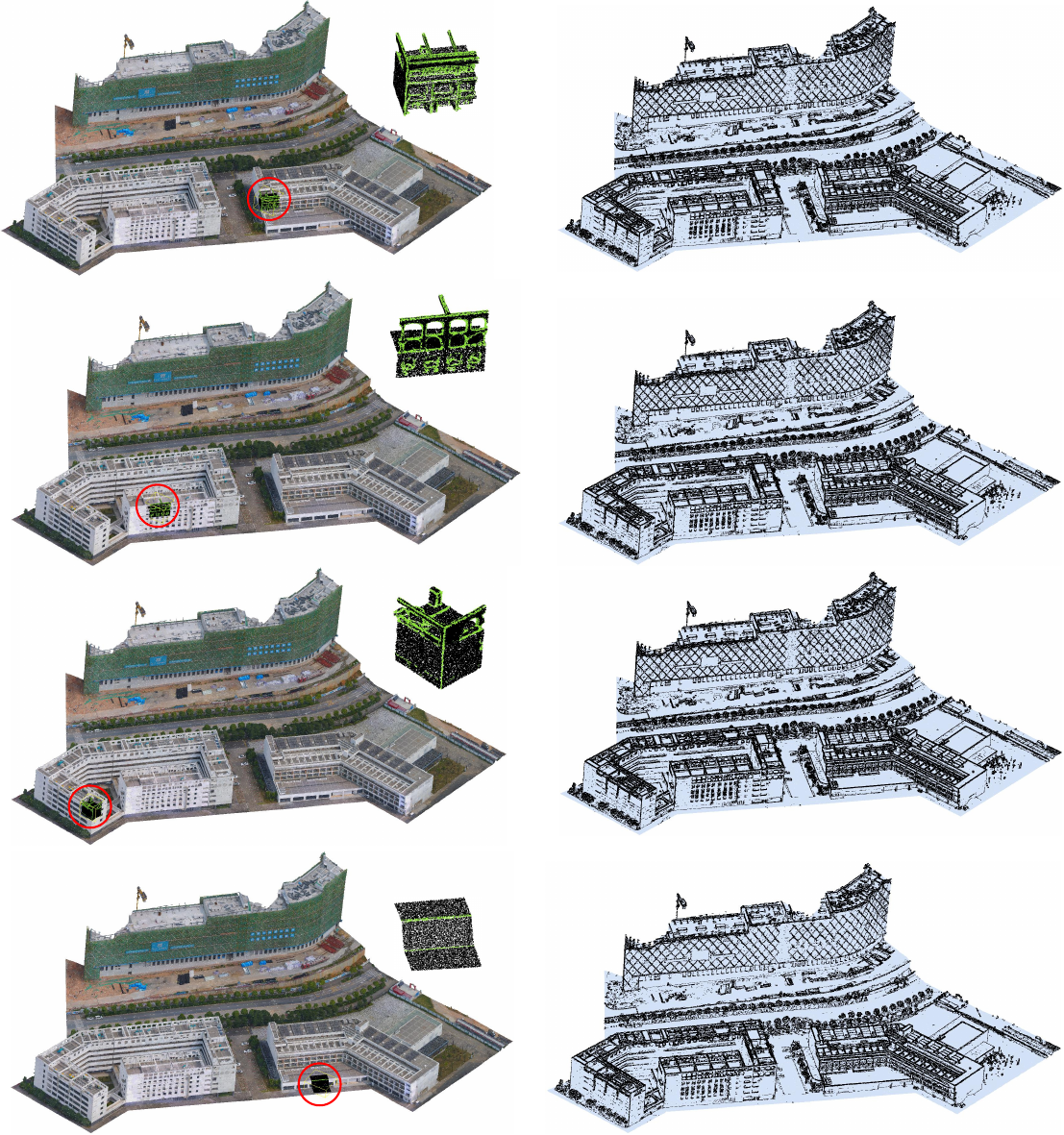}
  \caption{More results from different training scene segments on the UrbanBIS\cite{Yang2023UrbanBISAL} dataset. The green points and black points are training edges and non-edges, and the black points are the edges obtained by the proposed OSFENet. }
  \label{fig:segs}
\end{figure*}

\begin{figure*}
  \centering
  \includegraphics[width=\linewidth]{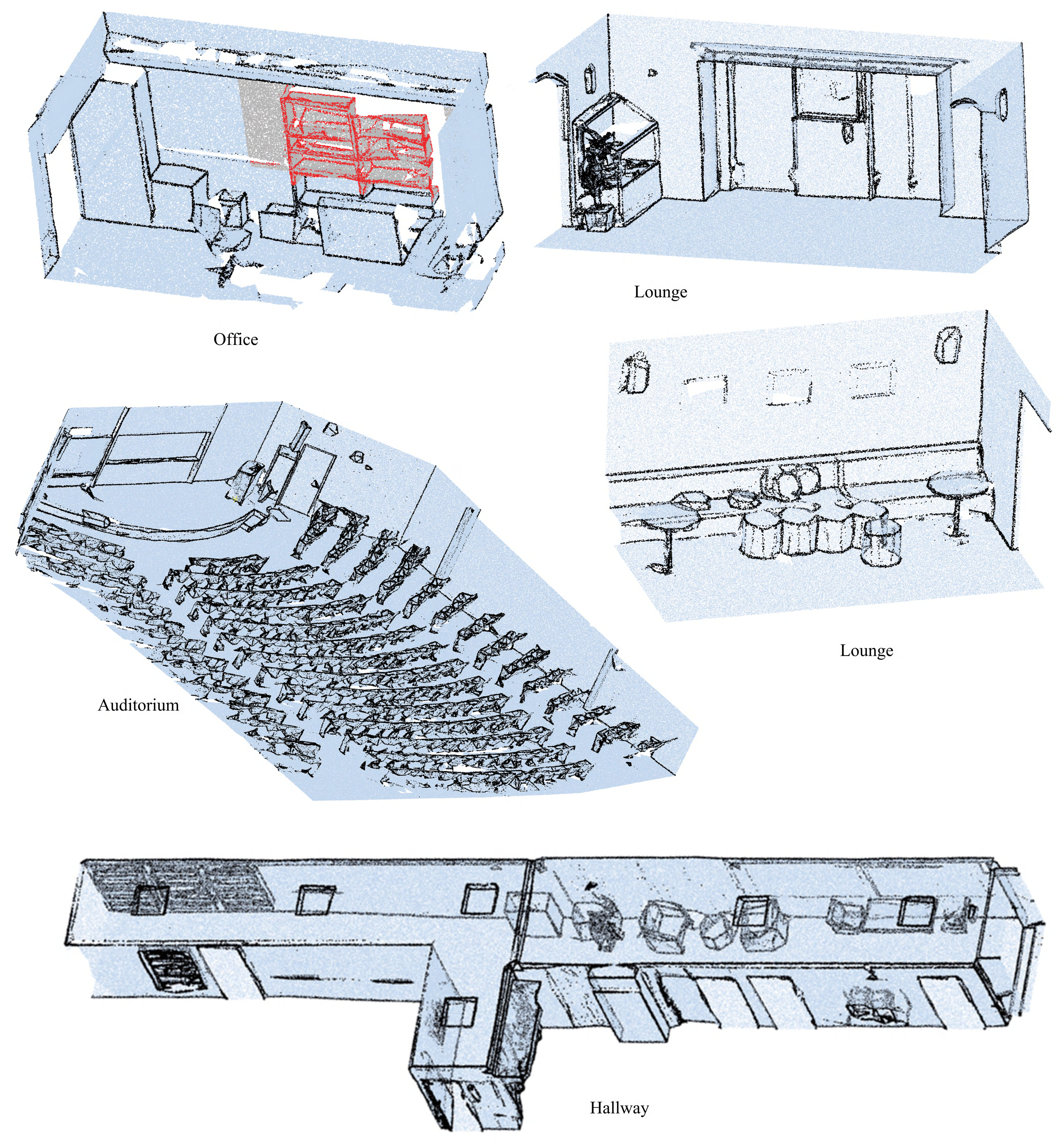}
  \caption{Results of edge detection on the S3DIS\cite{Armeni20163DSP} dataset. The red points and gray points are training edges and non-edges, and the black points are the predicted edges obtained by the proposed OSFENet.}
  \label{fig:indoor}
\end{figure*}

\begin{figure*}
  \centering
  \includegraphics[width=\linewidth]{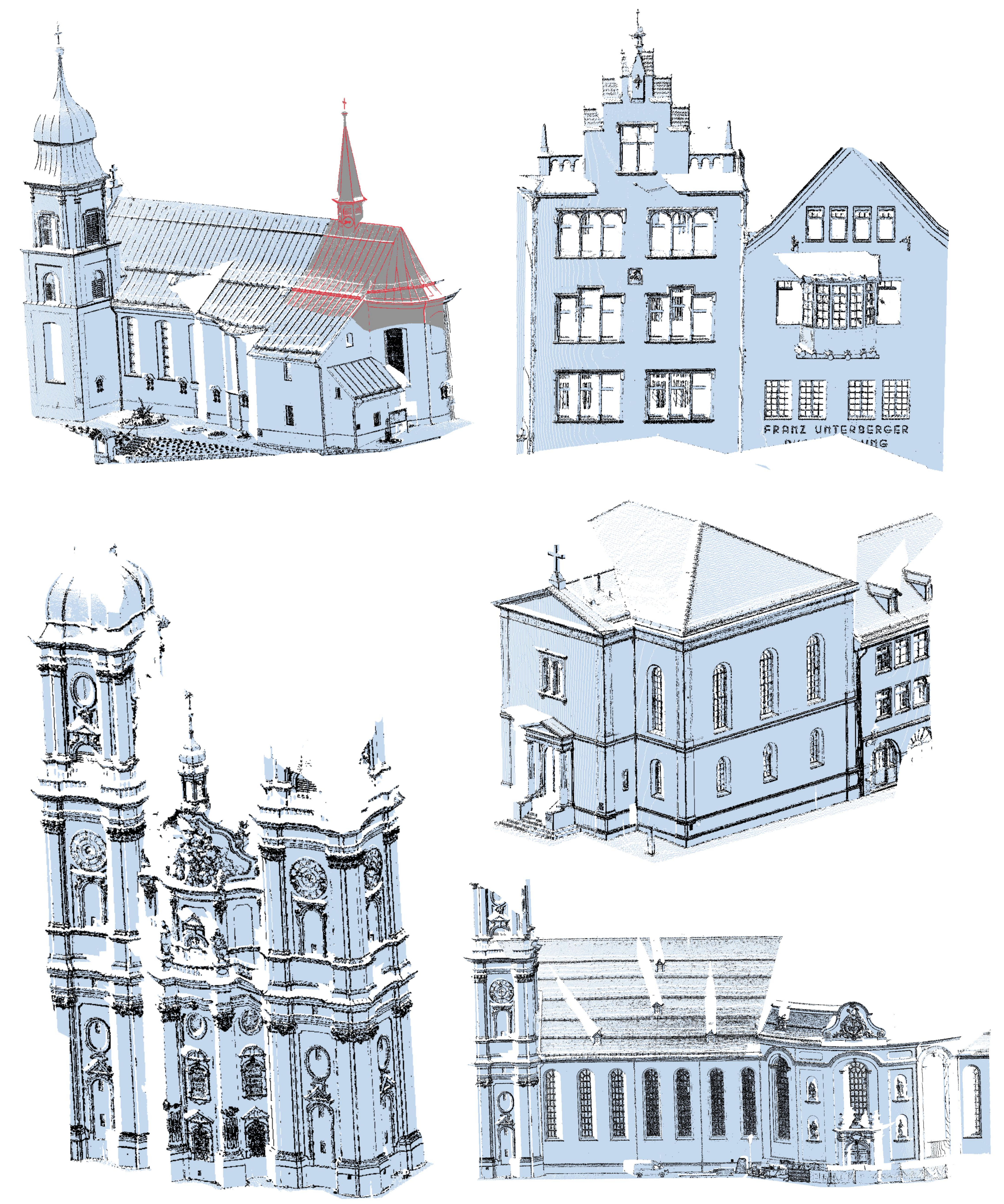}
  \caption{Results of edge detection on the Semantic3D\cite{Hackel2017Semantic3DnetAN} dataset. The red points and gray points are training edges and non-edges, and the black points are the predicted edges obtained by the proposed OSFENet.}
  \label{fig:Semantic3D}
\end{figure*}

\begin{figure*}
  \centering
  \includegraphics[width=\linewidth]{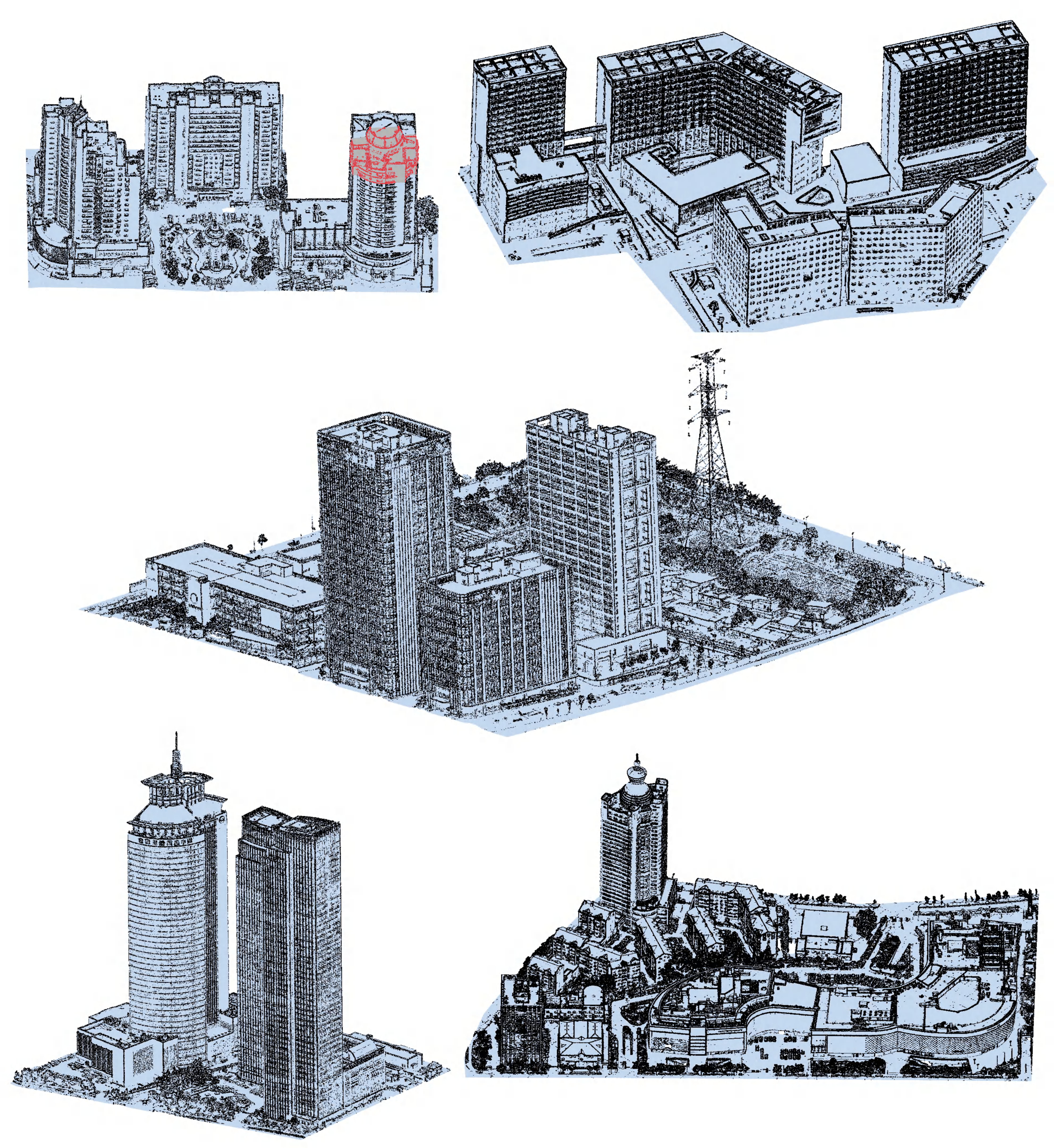}
  \caption{Results of edge detection on the UrbanBIS\cite{Yang2023UrbanBISAL} dataset. The red points and gray points are training edges and non-edges, and the black points are the predicted edges obtained by the proposed OSFENet.}
  \label{fig:UrbanBIS}
\end{figure*}

% \section{Additional results on real-scanned datasets}
% \begin{figure*}
%   \centering
%   \includegraphics[width=\linewidth]{Figure/ComparisonSupp.pdf}
%   \caption{Results on the real-scanned datasets, which are produced by OSFENet, EC-Net\cite{Yu2018ECNet} and NerVE\cite{Nerve2023CVPR}.}
%   \label{fig:comparison}
% \end{figure*}

% \subsection{Comparison with competitors}

% We choose EC-Net\cite{Yu2018ECNet} and NerVE\cite{Nerve2023CVPR} as baselines for comparison on real-scanned datasets, as they achieved the first and second highest scores on the ABC dataset. Figure \ref{fig:comparison} presents the testing results of our OSFENet, EC-Net\cite{Yu2018ECNet} and NerVE\cite{Nerve2023CVPR}, on the real-scanned datasets. 

% It can be observed that the EC-Net only extracts extremely sharp features, while it misses all the smooth ones. This is because EC-Net primarily focuses on sharp edges. Unlike the ABC dataset, indoor and outdoor scenes display a wider range of scales and diverse levels of detailed features, presenting additional challenges. Regarding the most recent method, NerVE, nearly all feature points are disregarded due to the normalization and voxelization operations. Achieving effective results becomes difficult when attempting to set a consistent voxel scale for models of various sizes, particularly for outdoor scenes. However, our OSFENet still performs better on these real-scanned datasets.

\section{More results from different training scene segments}

To examine the impact of training segment selection in real-scanned scenes on edge detection, a single scene from the UrbanBIS \cite{Yang2023UrbanBISAL} dataset was selected. Four smaller segments, representing both complex and simple features, were randomly selected and annotated. One segment was used for training, while the remaining point clouds served as test data. 

As illustrated in Figure \ref{fig:segs}, the results demonstrate that edges can be effectively detected regardless of whether the training is conducted on annotated segments containing simple or complex features. However, it is noteworthy that training on segments with only sharp edges limits the model to detecting sharp edges (as shown in the last example in Figure \ref{fig:segs}), whereas training on segments with diverse features enables the detection of a broader range of edges, including smoother ones.

\section{Additional results}
For a specific dataset, by learning the labeled segments of the scene, the proposed OSFENet can achieve satisfactory results for predicting the edges of the models from that dataset. Here, we present additional results on the aforementioned real-scanned datasets, including the indoor scene dataset S3DIS \cite{Armeni20163DSP}, as well as the outdoor scene datasets like Semantic3D \cite{Hackel2017Semantic3DnetAN} and UrbanBIS \cite{Yang2023UrbanBISAL}.

Indoor scenes and outdoor scenes differ in scale, shape, sampling density, and noise distribution. Furthermore, these two outdoor scenes exhibit edges with varying scales, shapes, and levels of detail. Specifically, the buildings in Semantic3D exhibit both sharp and smooth edges, whereas the buildings in UrbanBIS showcase more massive and detailed edges. Furthermore, the size of building instances varies in outdoor scenes. For example, in UrbanBIS, it ranges from less than 2m to more than 100m. 

As can be seen from Figure \ref{fig:indoor} to Figure \ref{fig:UrbanBIS}, the proposed OSFENet produces superior results on these real-scanned datasets with different characteristics.

\section{Results on models with holes}

\begin{figure}[H]
  \centering
  \includegraphics[width=\linewidth]{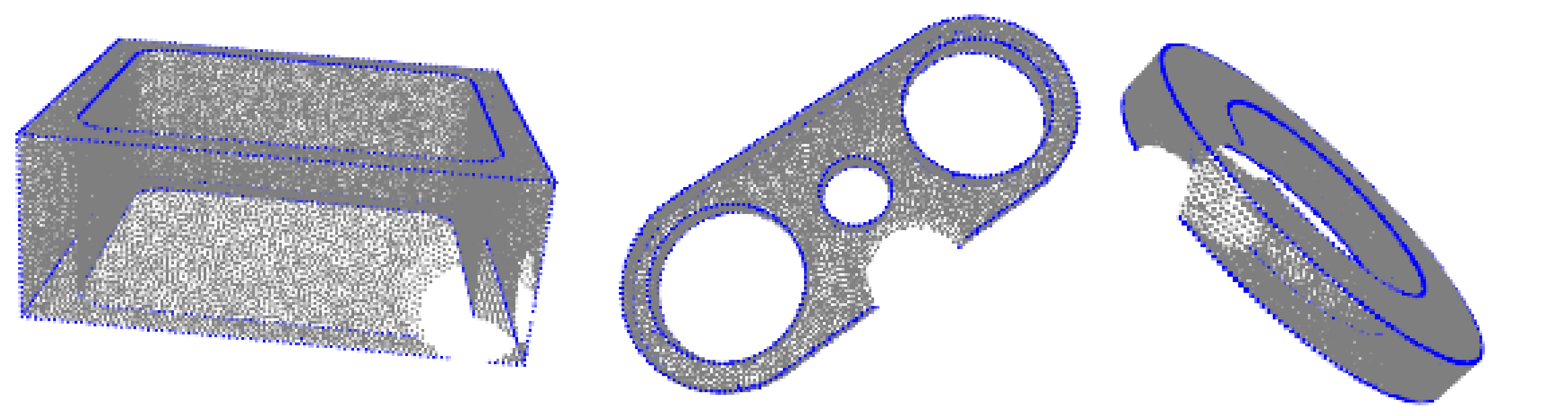}
  \caption{The results of our method on models with holes.}
  \label{fig:holes}
\end{figure}

We test our OSFENet on point clouds with holes, as shown in Figure \ref{fig:holes}, the point clouds are incomplete with several holes on the surfaces. However, the results indicate that when holes are located on smooth surfaces, no additional features are detected. Additionally, when the holes are situated on edges, the hole edges are missing because such edges are not considered features in the ground truth.

\section{Failure cases}

\begin{figure}[H]
  \centering
  \includegraphics[width=\linewidth]{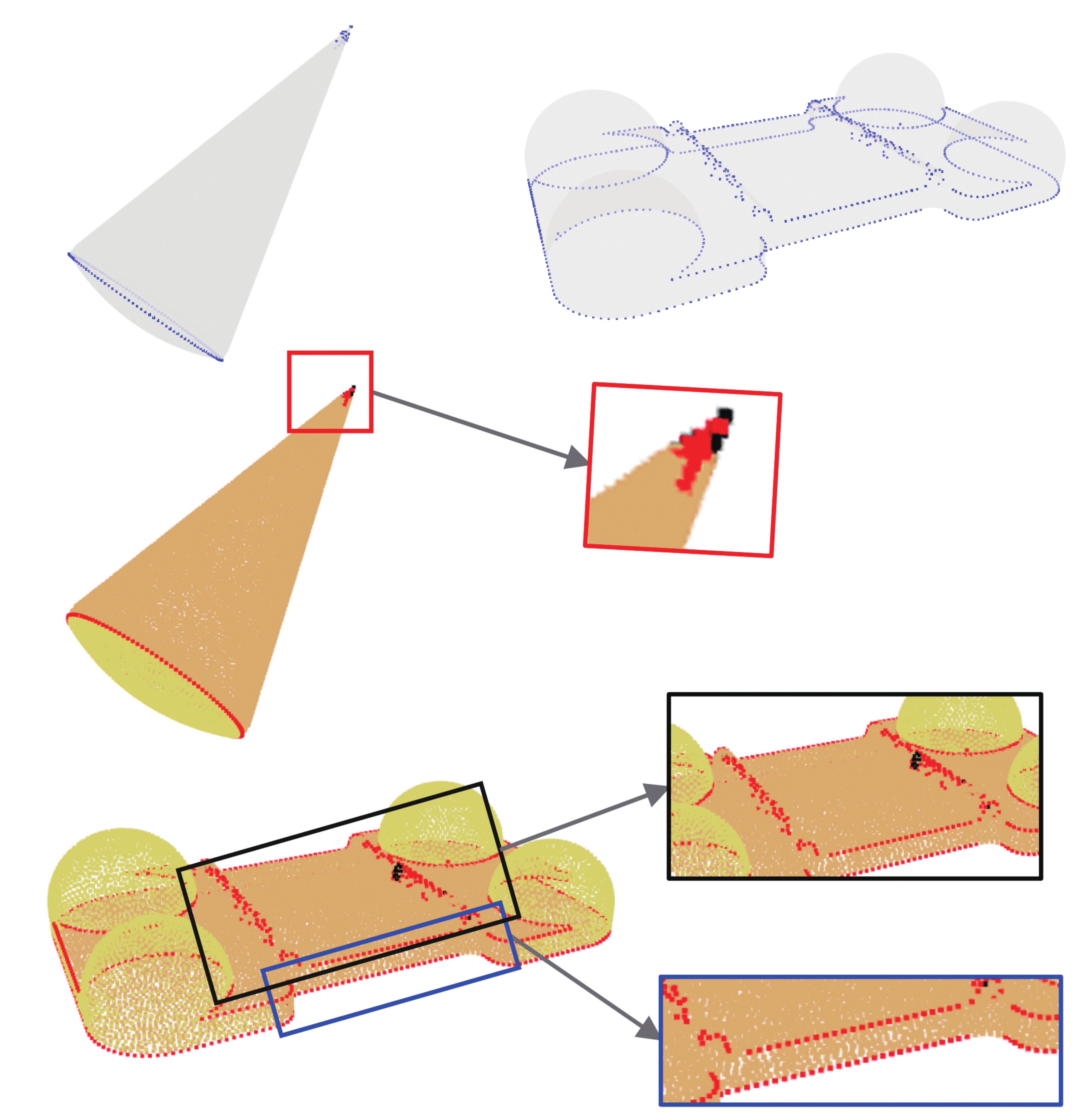}
  \caption{Two examples of failure cases in edge detection.}
  \label{fig:failurecase}
\end{figure}

Figure \ref{fig:failurecase} illustrates two failure cases involving edge detection and surface segmentation, while Table \ref{tab:failseg} provides the metrics scores for the corresponding surface segmentation. We observed that local patches generated using filtered $k$NN fail to account for the overall shape of the edges, causing our method to misinterpret apex regions or points on surfaces with subtle curvature as edge points. It would be interesting as future work to integrate both local and large-scale patches to account for the entire shape of the edges while simultaneously reducing the amount of training data required.

The surface segmentation using the straightforward flood-filling approach (BFS) is prone to failure in scenarios with small edge gaps (blue rectangle, Figure \ref{fig:failurecase}). Additionally, erroneous edge extraction can either segment additional surfaces (red rectangle) or leave results unchanged if the edges do not form a complete boundary (black rectangle).

% To enhance the performance of surface segmentation based on the detected edges, fitting edge curves to bridge these gaps could provide an effective solution.

% Moreover, it effectively filters out noise in grid-like regions of the scene.

% As our method can accurately distinguish the edge points, we can segment different surfaces of a model by using the predicted edge points as boundaries and performing the segmentation process with BFS (Broad First Search). Specifically, the $k$-NN graph ($k$ = 5) is constructed over the point cloud, the BFS starting from a random non-visited and non-edged point (called seed point), and stopping when encountering an edge point, is performed repeatedly until no seed point can be found. Some examples of surface segments are shown in Figure  \ref{fig:surface}, which further highlights the utility of the proposed method. 

% \begin{figure}[H]
%   \centering
%   \includegraphics[width=\linewidth]{Figure/surface.png}
%   \caption{The segmented surfaces.}
%   \label{fig:surface}
% \end{figure}

\section{Application}
Excluding surface segmentation, a wide range of applications that rely on features can benefit from this approach. We summarize the other applications in the following three areas: (i) Point cloud simplification: should retain both sharp and smooth features, large-scale and detailed elements, as well as structural and textural characteristics, while removing non-features. Therefore, our method can be employed as an up-stream technique for point cloud simplification; (ii) Model editing: refers to the process of modifying or altering a 3D model's geometry or appearance, for example, adjusting textures or materials, or refining details for a specific purpose. Therefore, our method can segment the model into finer grains, assisting the user in efficiently selecting elements, such as specific faces on a building. (3) Line drawing generation: is an important method of non-photorealistic rendering in computer graphics, as line drawings (also referred to as sketches) can better communicate main visual information. For example, from Figure\ref{fig:indoor} to Figure\ref{fig:UrbanBIS}, we can rapidly grasp the main content in each scene. Furthermore, line drawings can also serve as the foundation for further artistic creation.

% \section{Discussion on Limitations}

% The main idea is to learn the point distribution characteristic of a specific dataset to achieve distinguished performance compared to models trained on datasets compiled using different techniques. We have validated our method on both CAD model datasets and real-scanned datasets, and the superior results demonstrate its effectiveness and advantage. 

% However, to further validate the performance of our OSFENet, we also tested it on another object dataset, \textit{i.e.} SHREC, which consists of models acquired by different scanners and techniques. The training model and the predictions of our OSFENet, EC-Net\cite{Yu2018ECNet}, and NerVE\cite{Nerve2023CVPR} are presented in Figure \ref{fig:limitation}. As shown in Figure \ref{fig:limitation}, although some non-features in the vicinity of edges are erroneously taken as edge points, our method can still accurately extract the sharp features of the models. However, both EC-Net \cite{Yu2018ECNet} and NerVE \cite{Nerve2023CVPR} fail to extract sufficient sharp features describing the silhouettes of the models. 

\begin{table}[H]
  \caption{Failure cases of surface segmentation.}
  \label{tab:failseg}
  \small
  \begin{tabular}{llllll}
    \toprule
     {Models} &  {CD$\downarrow$}  &  {IoU$\uparrow$} &  {Precision$\uparrow$}  &  {Recall$\uparrow$} &  {F-score$\uparrow$} \\
    \midrule
     { Model1 } &  { 2.53e-5 } &  { 0.9976 } &  {1.00} &  {0.9976} &  {0.9988}  \\
     {Model2 } &  { 0.0105 } &  { 0.7414} &  {0.9924} &  {0.7456} &  {0.8515}  \\   
    \bottomrule
  \end{tabular}
\end{table}

\end{document}